%% file: example_paper.tex
\documentclass{article}

\usepackage{microtype}
\usepackage{graphicx}
\usepackage{subcaption}
\usepackage{booktabs} 

\usepackage{hyperref}
\usepackage{url}
\usepackage{stfloats}

\usepackage{graphicx} 
\usepackage{amsmath}
\usepackage{amssymb}
\usepackage{latexsym}
\usepackage{listings}
\usepackage{booktabs}
\usepackage{multirow}
\usepackage{wrapfig}
\usepackage{array}
\usepackage[most]{tcolorbox}
\usepackage{subcaption}
\usepackage{threeparttable}
\usepackage{tablefootnote}

\usepackage{pifont}
\usepackage{amsmath}  
\usepackage{amsfonts} 
\usepackage{xcolor}
\usepackage{colortbl}
\definecolor{skyblue}{RGB}{241, 246, 247}
\definecolor{lightgreen}{RGB}{244, 249, 243}
\definecolor{grey}{RGB}{242, 242, 242}
\definecolor{darkgreen}{RGB}{0, 142, 57}
\definecolor{darkred}{RGB}{212, 0, 0}
\definecolor{darkyellow}{RGB}{177, 160, 17}
\definecolor{darkgrey}{RGB}{110, 110, 110}

\usepackage[preprint]{icml2026}

\usepackage{amsmath}
\usepackage{amssymb}
\usepackage{mathtools}
\usepackage{amsthm}

\tcbuselibrary{theorems, skins, breakable} 

\usepackage[capitalize,noabbrev,nameinlink]{cleveref}

\theoremstyle{plain}

\theoremstyle{definition}

\theoremstyle{remark}

\usepackage[textsize=tiny]{todonotes}

\icmltitlerunning{BUDDY: \underline{BU}dget-\underline{D}riven \underline{DY}namic Depth Routing for Adaptive Large Language Model Inference}

\begin{document}

\newtcbtheorem[auto counter,
  crefname={Observation}{Observations},
  Crefname={Observation}{Observations}
]{observation}{Observation}{
  enhanced,
  colback=gray!8,
  colframe=black,
  boxrule=1pt,
  left=2mm, right=2mm, top=2mm, bottom=2mm,
  detach title,            
  attach title to upper,   
  after title={:\ },       
  coltitle=black,              
  fonttitle=\bfseries\itshape, 
  fontupper=\itshape,          
  breakable,
}{obs}

\twocolumn[
  \icmltitle{BUDDY: \underline{BU}dget-\underline{D}riven \underline{DY}namic Depth Routing for \\ Adaptive Large Language Models Inference}



  \icmlsetsymbol{equal}{*}

  \begin{icmlauthorlist}
    \icmlauthor{Yuhua Zhou}{zju,zjlab}
    \icmlauthor{Shaoqi Yu}{ant}
    \icmlauthor{Shichao Weng}{fdu}
    \icmlauthor{Changhai Zhou}{fdu}
    \icmlauthor{Mingze Yin}{zju}
    \icmlauthor{Fei Yang}{zjlab}
    \icmlauthor{Aimin Pan}{zjlab}
  \end{icmlauthorlist}

  \icmlaffiliation{zju}{College of Computer Science and Technology, Zhejiang University}
  \icmlaffiliation{zjlab}{Zhejiang Lab}
  \icmlaffiliation{fdu}{School of Computer Science, Fudan University}
  \icmlaffiliation{ant}{Ant Group}

  \icmlcorrespondingauthor{Aimin Pan}{panaimin@zhejianglab.org}
   \icmlcorrespondingauthor{Fei Yang}{yangf@zhejianglab.org}

  \icmlkeywords{Machine Learning, ICML}

  \vskip 0.3in
]



\printAffiliationsAndNotice{}  

\begin{abstract}

Large language models (LLMs) incur high inference cost due to their depth and parameter scale. Depth pruning can reduce latency by skipping redundant Transformer blocks, but existing methods (i) provide limited control under user-specific compute budgets and (ii) typically fix the routing path, failing to adapt as the context grows during decoding. We propose \textsc{Buddy}, a budget-driven dynamic depth routing framework. \textsc{Buddy} uses a lightweight Decision Module to score intermediate layers conditioned on the input and deterministically executes the top-$k$ layers to satisfy a given budget. To support decode-time adaptation, \textsc{Buddy} reuses the first-layer KV cache as a low-overhead global context source and pools it together with the newest token representation before each routing decision. When no explicit budget is provided,an optional Budget Predictor estimates an input-dependent compute level to balance quality and efficiency. Experiments on Llama-family and Qwen models show that \textsc{Buddy} is competitive with strong static pruning baselines and often improves the accuracy–compute trade-off, while uniquely supporting strict budget control, decode-time rerouting, and multiple budgets within a single trained model.

\end{abstract}

\input{pages/01_intro}

\input{pages/02_related}
\input{pages/03_background}

\input{pages/04_method}

\input{pages/05_evaluation}
\input{pages/06_conclusion}

\bibliography{example_paper}
\bibliographystyle{icml2026}

\newpage
\appendix
\onecolumn
\input{pages/appendix}

\end{document}

%% file: pages/01_intro.tex
\section{Introduction}\label{sec:intro}

Large language models (LLMs) achieve strong performance on a wide range of NLP tasks \citep{chang2024survey,zhou2026lara}, yet their inference latency remains high because computation scales with model depth and width \citep{zhou2025bslora,jiang2024artfl}. A practical way to reduce latency is depth pruning, which skips entire Transformer blocks without changing tensor and attention shapes. 

Most existing depth-pruning methods are \emph{static}: they estimate a global importance score for each layer and permanently remove a fixed subset prior to deployment \citep{kim2024shortened,song2024sleb}. 
Static pruning is simple and highly efficient at runtime because of its fixed execution graph but cannot adjust its execution path to input difficulty or heterogeneous user budgets \citep{wee2025pudding}.
Recent \emph{dynamic} approaches improve flexibility by skipping layers on the fly (e.g., via routing \citep{jiang2024dllm} or early-exit heuristics \citep{fan2024notall,elhoushi2024layerskip}). However, they typically: (1) enforce a \textbf{fixed sparsity} pattern or weakly control the number of executed layers, which makes the actual computation unpredictable; and (2) determine a \textbf{fixed routing path} during the prefill stage and reuse it during decoding, even though the importance of layers can change as new tokens are generated.

In this paper, we seek an inference mechanism that \textbf{adapts model depth to both input difficulty and user budgets}, so that \textbf{a single deployed LLM can reliably serve heterogeneous compute constraints.} This setting raises two challenges. 
\textbf{(C1) Input-adaptive routing:} layer importance is highly prompt-dependent (Observation \ref{obs:observation_1}), not only across but also within tasks, requiring per-input decisions to preserve quality under a strict budget. 
\textbf{(C2) Decode-adaptive routing:} in autoregressive decoding, the relative utility of layers can shift as new tokens extend the context (Observation \ref{obs:observation_2}). Yet decode-time routing is difficult because each step introduces only one new token, offering a weak local signal unless the router can access a compact summary of the full prefix. Addressing \textbf{(C2)} therefore requires injecting global context into routing decisions at every step without negating the compute savings.

\begin{table*}[t]
\centering
\caption{
A holistic comparison of \textsc{Buddy} with related approaches: static depth pruning and dynamic depth routing. \textsc{Buddy}supports all four properties simultaneously under this definition. 
\textbf{Budget Strict:} deterministic control over the exact number of executed Transformer blocks for the user's budget constraint. 
\textbf{Budget Flexibility:} the same model meets users' diverse budgets. 
\textbf{Input Adaptive:} automatically adapts the inference path based on the input. 
\textbf{Decode Adaptive:} dynamically adjusting its decoding path during the generation process.
}
\label{tab:teaser_table}
\resizebox{\hsize}{!}{
\begin{tabular}{l|c|cccc}
\toprule
Method & Pruning 
& Budget Strict 
& Budget Flexibility 
& Input Adaptive 
& Decode Adaptive 
\\

\midrule
Shortened Llama \citep{kim2024shortened} & \textcolor{darkred}{static} & \color{darkgreen}\ding{51} & \color{darkred}\ding{55} & \color{darkred}\ding{55} & \color{darkred}\ding{55} \\
ShortGPT \citep{men2024shortgpt}         & \textcolor{darkred}{static} & \color{darkgreen}\ding{51} & \color{darkred}\ding{55} & \color{darkred}\ding{55} & \color{darkred}\ding{55} \\
SLEB \citep{song2024sleb}                & \textcolor{darkred}{static} & \color{darkgreen}\ding{51} & \color{darkred}\ding{55} & \color{darkred}\ding{55} & \color{darkred}\ding{55} \\
\midrule
Early-exit \citep{elhoushi2024layerskip,fan2024notall} & \textcolor{darkgreen}{dynamic} & \color{darkred}\ding{55} & \color{darkred}\ding{55} & \color{darkgreen}\ding{51} & \color{darkred}\ding{55} \\
Token-wise \citep{yang2025dash,jiang2024dllm} & \textcolor{darkgreen}{dynamic} & \color{darkred}\ding{55} & \color{darkred}\ding{55} & \color{darkgreen}\ding{51} & \color{darkgreen}\ding{51} \\
PuDDing \citep{wee2025pudding}           & \textcolor{darkgreen}{dynamic} & \color{darkgreen}\ding{51} & \color{darkred}\ding{55} & \color{darkgreen}\ding{51} & \color{darkred}\ding{55} \\
FiRST \citep{jain2025first}              & \textcolor{darkgreen}{dynamic} & \color{darkred}\ding{55} & \color{darkred}\ding{55} & \color{darkgreen}\ding{51} & \color{darkred}\ding{55} \\
\midrule
\textbf{\textsc{Buddy} (Ours)}                    & \textcolor{darkgreen}{dynamic} & \color{darkgreen}\ding{51} & \color{darkgreen}\ding{51} & \color{darkgreen}\ding{51} & \color{darkgreen}\ding{51} \\
\bottomrule
\end{tabular}}
\end{table*}

We propose \textbf{\textsc{Buddy}}, a \textbf{\underline{BU}}dget-\textbf{\underline{D}}riven \textbf{\underline{DY}}namic depth routing framework for LLMs. Given an input and an optional compute budget, \textsc{Buddy} selects a subset of layers to execute that (a) satisfies the budget constraint, and (b) preserves task performance. Concretely, to address the first challenge, \textsc{Buddy} introduces a lightweight \emph{Decision Module} that scores intermediate layers conditioned on the current context and selects the top-$k$ layers to execute, where $k$ is tied to the user budget. To stabilize learning and accelerate adaptation, we optionally initialize the Decision Module with \emph{static priors} derived from standard importance indicators (e.g., perplexity-, Taylor-, or representation-similarity–based signals) after appropriate normalization and fusion.

To address the second challenge, \textsc{Buddy} introduces a mechanism to inject \emph{global information during decoding}: we reuse the first layer’s KV cache to summarize the global context and concatenate the newest token’s features before each routing decision. This provides stable, low-overhead global signals to the Decision Module and enables \textsc{Buddy} to update the execution path as the generation unfolds.

Furthermore, when users omit a budget, \textsc{Buddy} turns to an \emph{Adaptive Budget Predictor} that provides the optimal budget according to the inputs. In our experiments on the Llama family, \textsc{Buddy} achieves stronger accuracy–compute trade-offs than static and dynamic baselines at matched sparsity, while providing deterministic control over the executed compute.

We emphasize that  \textsc{Buddy} targets a different serving setting: a single deployed model must satisfy diverse user budgets, preserve strict control over executed depth, and adapt its routing path as the input and decoding context evolve. A concise summary appears in \Cref{tab:teaser_table}. We therefore evaluate BUDDY as an accuracy–efficiency–flexibility trade-off rather than as a replacement for all static pruning regimes. Our contributions are summarized as:
\begin{itemize}
  \item \textbf{Budget-aware decision making.} We design a lightweight Decision Module that scores intermediate layers conditioned on the current context and selects the top-$k$ layers to execute, providing \emph{explicit} and \emph{deterministic} control over compute under user-specified budgets.
  \item \textbf{Global information during decoding.} We introduce a low-overhead scheme that reuses the first layer’s KV cache to expose global context to the router at every decoding step, enabling path updates as generation evolves.
  \item \textbf{Adaptive budget prediction.} When users do not specify a budget, a discrete Budget Predictor chooses a compute level that preserves task quality while minimizing executed layers.
  \item \textbf{Extensive empirical study.} Across Llama-family and Qwen models and multiple benchmarks, \textsc{Buddy} achieves competitive or best average accuracy in most settings while supporting strict budget control, decode-time adaptation, and multiple sparsity levels with a single trained model.
\end{itemize}

%% file: pages/02_related.tex
\section{Related Work}\label{sec:related_work}


\subsection{Static Depth Pruning}
Depth pruning removes redundant Transformer blocks to reduce inference latency and memory \citep{wang2024model}.  Recent studies typically score layer importance and drop low-scoring blocks. Methods such as ShortGPT \citep{men2024shortgpt} and Shortened LLaMA \citep{kim2024shortened}  rank blocks by Block Influence (cosine), Taylor score, or $\Delta\text{PPL}$. Iterative schemes such as SLEB \citep{song2024sleb} further identify redundancy and eliminate low-impact layers progressively. While effective, static pruning is task-agnostic and is fixed post-calibration, and cannot adapt to input difficulty or enforce user-specified compute budgets. Consequently, developing frameworks that are inherently compute-efficient yet robust to environmental constraints remains a high priority across diverse machine learning paradigms \citep{jiang2025towards}.

\subsection{Dynamic Depth Pruning}
Dynamic methods \citep{jiang2026dual} decide at inference which layers to execute based on the current input or context. \emph{Early-exit} methods  \citep{fan2024notall,elhoushi2024layerskip} produce predictions from intermediate layers and skip the rest, but may degrade quality on difficult instances by discarding later computations. \emph{Layer-skipping} methods aim to preserve the depth of reasoning while avoiding redundant blocks via designing routers to skip computation based on token-level \citep{jiang2024dllm,raposo2024mixture,yang2025dash,luo2025flexidepth} or prompt-level \citep{wee2025pudding,jain2025first} signals. Furthermore, PuDDing \citep{wee2025pudding} trains a prompt-conditioned global router that selects an omission set of blocks before decoding. However, many existing dynamic pruning methods cannot strictly conform to the budget and must train separate models for different sparsity targets. Our work exposes budget as a first-class constraint and learning a router that honors user-specified budgets while updating paths throughout decoding.

%% file: pages/03_background.tex
\section{Background and Motivation}\label{backgounr_and_motivation}

\subsection{Background}
\paragraph{Layer Skipping. }\label{sec:layer_skipping}
Dynamic depth reduction accelerates inference by conditionally bypassing Transformer blocks while preserving residual alignment. Let $\mathcal{M}\in\{0,1\}^{L}$ be a binary skipping mask, where $\mathcal{M}_\ell=1$ means ``execute'' and $\mathcal{M}_\ell=0$ means ``skip.'' With hidden states $\{\mathcal{H}_\ell\}_{\ell=0}^{L}$ (and $\mathcal{H}_0$ the input to the stack), the residual update is: 
\begin{equation}
\mathcal{H}_\ell =
\begin{cases}
\mathcal{F}_{\ell}(\mathcal{H}_{\ell-1}) + \mathcal{H}_{\ell-1}, & \text{if } \mathcal{M}_{\ell}=1,\\[2pt]
\mathcal{H}_{\ell-1}, & \text{if } \mathcal{M}_{\ell}=0.
\end{cases}
\end{equation}

\subsection{Motivation}

\begin{observation}{}{observation_1}
The importance of Transformer layers is \emph{input-dependent}; different inputs induce different importance distributions across layers.
\end{observation}

\begin{figure}[t]
\centering
\includegraphics[width=\linewidth]{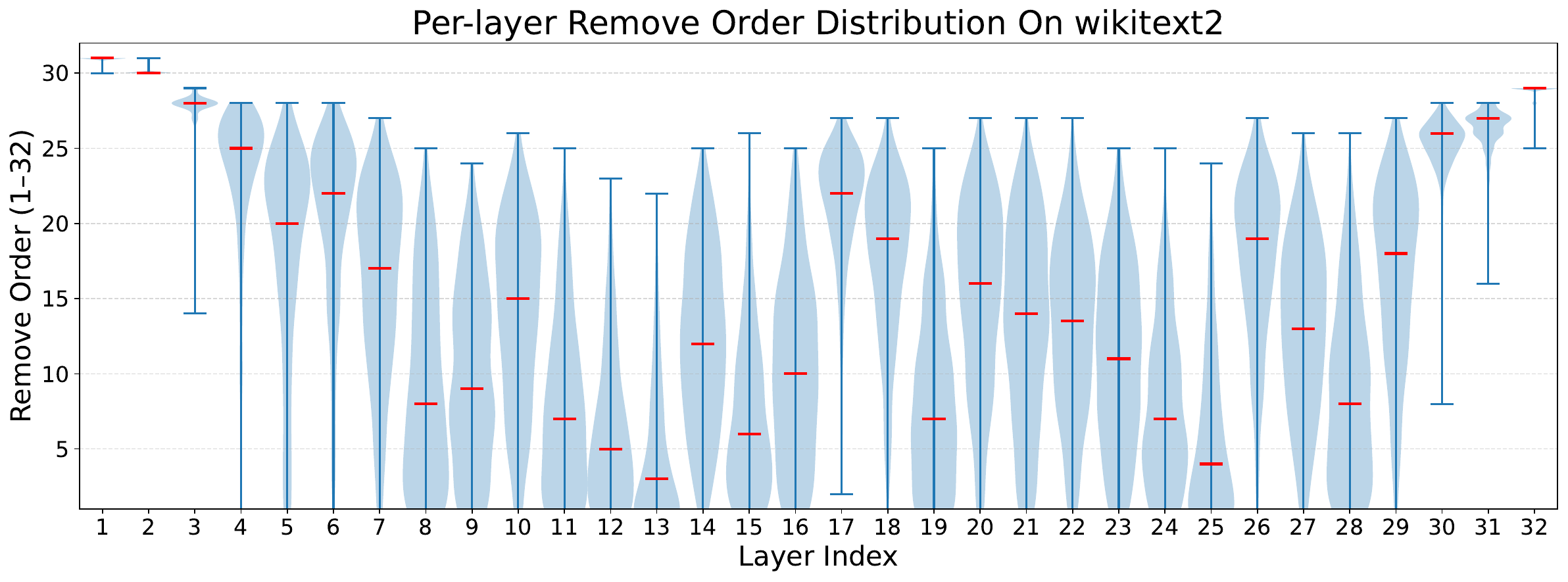}
\caption{
Input-dependent layer importance ranking distributions across different inputs on the WikiText-2 datasets. The chart shows how layer rankings vary significantly across inputs, suggesting the need for dynamic pruning decisions. 
}
\label{fig:observation_1_2}
\end{figure}

\paragraph{\textbf{Input-Adaptive. }}
We compute per-layer importance of Llama2-7B on WikiText-2 \citep{merity2017pointer} using the $\Delta\mathrm{PPL}$ score and, for each input, sort layers to obtain a \emph{remove order} (layers ranked earlier are less important and thus more removable). As illustrated in \Cref{fig:observation_1_2}, the resulting rank distributions vary markedly across inputs. For example, the average importance of layer 10 is 15th, while the importance range is relatively large (from 0th to 26th), and the importance distribution is not concentrated, indicating strong input dependence. This variability implies that the importance estimated once on a small validation set (static pruning) cannot accommodate diverse inputs; instead, pruning decisions should be made dynamically to adapt to different inputs.

\begin{observation}{}{observation_2}
The importance distribution \emph{evolves during decoding}; as generation proceeds, the relative utility of layers changes with the growing context.
\end{observation}

\begin{figure}[t]
\centering
\includegraphics[width=\linewidth]{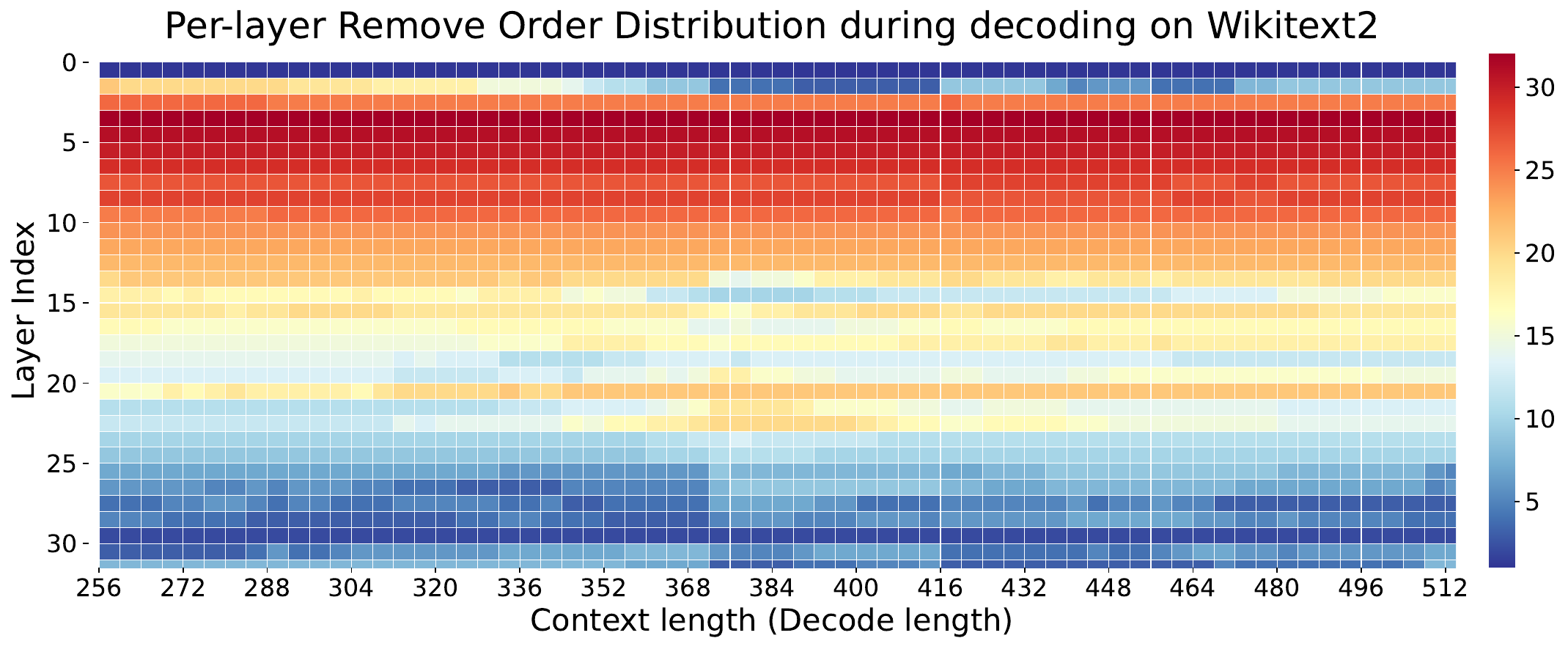}
\caption{
Evolution of layer importance ranking shifts with context length on the WikiText-2 dataset during autoregressive decoding. The heatmap illustrates how layer importance (e.g., layer 14) fluctuates throughout the decoding process.
}
\label{fig:observation_2}
\end{figure}    

\paragraph{\textbf{Decode-Adaptive. }}
To emulate different stages of autoregressive generation, we vary the context length from $256$ to $512$ and recompute layer importance. \Cref{fig:observation_2} shows that as context length increases, the per-layer ranking of Llama2-7B shifts substantially. For example, the importance of the 14th layer of the model is relatively important at the beginning, then becomes unimportant, and then becomes important again during the decoding process. Hence, a routing path fixed at prefill is insufficient; pruning must adapt throughout decoding as new tokens extend the context.

Based on the above two observations, we propose a dynamic layer execution method that can adapt to different inputs and select different layers for execution, while also dynamically modifying the inference path during the inference process.

%% file: pages/04_method.tex
\section{Method}\label{sec:method}

\begin{figure*}
\centering
\includegraphics[width=\linewidth]{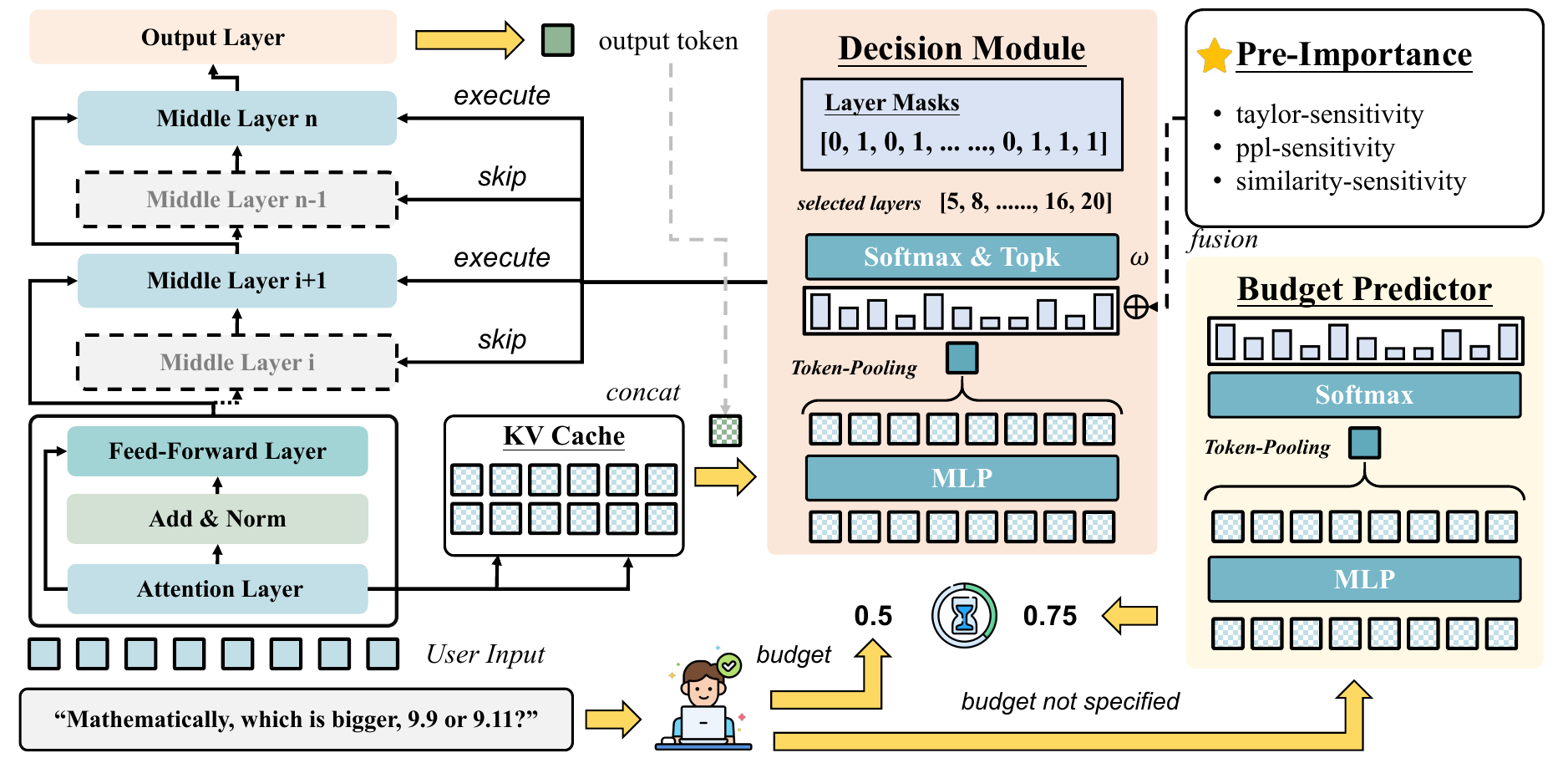}
\caption{
Overview of the \textsc{Buddy} framework. The framework consists of three key components: \textbf{(1) Decision Module} that adaptively selects layers based on context and budget constraints, \textbf{(2) KV-aware Planner} that reuses first-layer KV cache and updates global context during decoding, and \textbf{(3) Budget Predictor} that automatically provides estimated budgets when not explicitly provided.
}
\label{fig:overview}
\end{figure*}

\subsection{Framework Overview}
The framework overview of \textsc{Buddy} is summarized in \Cref{fig:overview}. Given a prompt $x$ and an optional compute budget $b$, \textsc{Buddy} selects a subset of Transformer blocks to execute. (1) It extracts a KV-aware \emph{global context signal} from the first layer (\Cref{sec:kv_aware}) and feeds it to a lightweight Decision Module (\Cref{sec:decision_module}) that outputs a score $s_\ell$ for each candidate middle block. (2) It deterministically executes the top-$k$ middle blocks, where $k$ is derived from $b$, yielding a binary mask $M\in\{0,1\}^{L_{\mathrm{mid}}}$ for inference. During autoregressive decoding, the global context is refreshed with the newest token, enabling decode-time re-routing. (3) When no budget is specified, a Budget Predictor (\Cref{sec:budget_predictor}) infers a discrete compute level $\hat{b}$ from the same context to balance quality and efficiency.

\begin{table*}[ht]
\centering
\caption{
Performance comparison on Llama3-8B across eight Commonsense reasoning benchmarks for different pruning methods at various sparsity levels (12.5\%-50\%). \textbf{Boldface} indicates the best performance, and the \underline{Underline} means the second-order performance. \textsc{Buddy} achieves the best or competitive average performance across pruning ratios while using a single multi-budget model.
}
\label{tab:result_llama8b}
\resizebox{\textwidth}{!}{
\begin{tabular}{l|c|c|ccccccccc}
\toprule
Method & Pruning & RM Blocks & OBQA & PIQA & BoolQ & SIQA & Hellaswag & ARC-E & ARC-C & Winogrande & Avg. \\
\midrule
Dense w/o & - & 0(0.0\%) & 44.80 & 80.85 & 80.98 & 46.98 & 79.17 & 80.09 & 53.24 & 73.40 & 67.44 \\
\midrule
Shortened Llama & static  & 4(12.5\%) & 43.40 & 79.27 & 73.06 & 47.34 & 75.32 & 76.85 & 48.04 & 66.77 & 63.76 \\
ShortGPT        & static  & 4(12.5\%) & \underline{44.20} & \underline{79.76} & \underline{79.48} & 48.16 & \textbf{76.99} & \textbf{79.88} & \underline{51.02} & \underline{73.72} & \underline{66.65} \\
SLEB            & static  & 4(12.5\%) & \underline{44.20} & \textbf{80.47} & 75.14 & \textbf{49.28} & \underline{76.69} & \textbf{79.88} & 50.77 & 72.22 & 66.08 \\
PuDDing         & dynamic & 4(12.5\%) & 38.80 & 76.82 & 68.53 & 44.83 & 72.93 & 74.12 & 44.45 & 71.98 & 61.56 \\
FiRST           & dynamic & 4(12.5\%) & 31.60 & 52.88 & 63.58 & 37.26 & 42.01 & 42.34 & 30.80 & 52.57 & 44.13 \\
\rowcolor{gray!12}\textsc{Buddy}           & dynamic & 4(12.5\%) & \textbf{44.80} & 79.71 & \textbf{83.24} & \underline{48.41} & 76.25 & \underline{78.96} & \textbf{53.33} & \textbf{74.51} & \textbf{67.40} \\
\midrule
Shortened Llama & static  & 8(25\%) & \underline{39.20} & \textbf{77.64} & 58.29 & 44.88 & \textbf{69.26} & \textbf{72.01} & 41.47 & 58.33 & 57.63 \\
ShortGPT        & static  & 8(25\%) & 39.00 & 73.67 & 64.25 & \underline{44.93} & \underline{69.23} & 70.96 & \textbf{47.44} & \textbf{71.74} & \underline{60.15} \\
SLEB            & static  & 8(25\%) & \textbf{39.40} & \textbf{77.64} & 55.32 & 44.58 & 69.17 & \underline{71.68} & 41.81 & 58.56 & 57.27 \\
PuDDing         & dynamic & 8(25\%) & 31.40 & 71.06 & 62.54 & 41.04 & 57.65 & 62.04 & 34.22 & 59.59 & 52.44 \\
FiRST           & dynamic & 8(25\%) & 33.00 & 55.33 & \underline{64.56} & 38.02 & 46.50 & 45.08 & 31.57 & 53.83 & 45.99 \\
\rowcolor{gray!12}\textsc{Buddy}           & dynamic & 8(25\%) & 39.00 & \underline{74.05} & \textbf{75.26} & \textbf{47.19} & 67.29 & 70.75 & \underline{45.99} & \underline{70.40} & \textbf{61.24} \\
\midrule
Shortened Llama & static  & 12(37.5\%) & \underline{35.00} & \textbf{72.69} & 61.80 & \underline{42.68} & \textbf{59.37} & \underline{63.76} & 33.45 & 57.54 & 53.29 \\
ShortGPT        & static  & 12(37.5\%) & 31.60 & 68.93 & \textbf{70.24} & \textbf{43.04} & \underline{58.97} & 56.14 & \textbf{38.40} & \textbf{67.56} & \underline{54.36} \\
SLEB            & static  & 12(37.5\%) & \textbf{35.80} & \underline{72.25} & 60.52 & 42.43 & 58.78 & \textbf{65.07} & 35.92 & 56.04 & 53.35 \\
PuDDing         & dynamic & 12(37.5\%) & 27.00 & 64.64 & 61.93 & 38.38 & 43.13 & 47.31 & 30.46 & 57.85 & 46.34 \\
FiRST           & dynamic & 12(37.5\%) & 33.60 & 55.71 & 64.46 & 37.87 & 45.00 & 43.60 & 31.66 & 52.64 & 45.57 \\
\rowcolor{gray!12}\textsc{Buddy}           & dynamic & 12(37.5\%) & 34.20 & 69.97 & \underline{68.96} & 42.32 & 58.76 & 63.05 & \underline{37.03} & \underline{60.93} & \textbf{54.40} \\
\midrule
Shortened Llama & static  & 16(50\%) & 28.20 & 66.00 & 55.29 & 38.95 & 43.89 & \underline{54.76} & 28.33 & 53.28 & 46.08 \\
ShortGPT        & static  & 16(50\%) & 28.40 & 65.29 & \textbf{65.20} & \underline{40.99} & \textbf{48.16} & 46.51 & 29.01 & \textbf{60.69} & \underline{48.03} \\
SLEB            & static  & 16(50\%) & \underline{30.20} & \textbf{68.50} & 53.00 & \textbf{41.86} & 45.04 & \textbf{56.06} & 27.47 & 52.57 & 46.84 \\
PuDDing         & dynamic & 16(50\%) & 23.40 & 58.60 & 52.97 & 36.08 & 31.92 & 34.76 & 23.72 & 48.86 & 38.79 \\
FiRST           & dynamic & 16(50\%) & \textbf{32.40} & 54.13 & \underline{62.60} & 37.77 & 42.80 & 42.17 & \textbf{30.20} & 52.80 & 44.36 \\
\rowcolor{gray!12}\textsc{Buddy}           & dynamic & 16(50\%) & 30.00 & \underline{67.68} & 61.68 & 39.61 & \underline{47.79} & 53.83 & \underline{30.12} & \underline{54.14} & \textbf{48.11} \\
\bottomrule
\end{tabular}}
\end{table*}

\subsection{Decision Module}\label{sec:decision_module}
\subsubsection{Formulation}\label{sec:decision_module_formulation}
The Decision Module selects which Transformer blocks to execute for a given input and budget. For an LLM with $L$ layers $\mathcal{B}=\{B_1,\ldots,B_L\}$, each inference step corresponds to a binary mask over layers, yielding $2^L$ possible execution paths. Exhaustive search is intractable, so we adopt a \emph{ranking-and-selection} formulation: we estimate per-layer importance conditioned on the current context and execute the top-$k$ layers determined by the budget. 

Following prior work \citep{jiang2024dllm}, we always execute the first and last blocks. Let $\mathcal{B}_{\mathrm{mid}}=\{B_2,\ldots,B_{L-1}\}$ denote the $L_{\mathrm{mid}}=L-2$ candidate middle layers. Given a context representation $H\in\mathbb{R}^{n\times D}$, a lightweight scorer $F(\cdot)$ produces scores $s\in\mathbb{R}^{L_{\mathrm{mid}}}$. We optionally fuse $s$ with a normalized offline prior $\tilde p$ (\Cref{sec:decision_module_Prior}) and obtain routing probabilities via $\mathrm{Softmax}$. We define the budget $b \in (0,1]$ as the target fraction of executed layers. The $b$ is converted to an integer $k=\mathrm{round}(b\cdot L)-2$, and we execute the $k$ highest-scoring middle layers. This produces a deterministic mask $M\in\{0,1\}^{L_{\mathrm{mid}}}$ that gates the residual updates of the middle blocks. The computation is formulated as: 

\vspace{-10pt}
\begin{equation}
\label{eq:formulation}
\resizebox{0.9\linewidth}{!}{$
\begin{gathered}
    s' = s + \omega\,\tilde p, \qquad \tilde p = \frac{p - \mu(p)}{\sigma(p) + \varepsilon} \\[6pt]
    \mathcal{P}(s', b) = \mathrm{Top}\text{-}K\big(\mathrm{Softmax}(s'),\,k = \mathrm{round}(b \cdot L) - 2\big)
\end{gathered}
$}
\end{equation}
where a small $\omega$ balances model-predicted scores and the prior. The middle layers then perform a single forward pass with dynamic depth. Layer selection during inference can be viewed as a binary decision per layer (execute vs.\ skip): blocks with $m_\ell=1$ are executed; blocks with $m_\ell=0$ are skipped via identity routing while preserving residual alignment. 

\subsubsection{Prior Normalization and Fusion}\label{sec:decision_module_Prior}
We compute offline layer priors on a small validation set (e.g., WikiText2 and PTB), following prior work on depth pruning \citep{kim2024shortened}. We consider three types of indicators to initialize the prior (e.g., $\Delta\text{PPL}$ when skipping layer $\ell$, Taylor/Fisher scores, and cosine-based dissimilarity). Specifically, we use one prior indicator at a time rather than jointly combining all three. Because different indicators have different scales and units, we normalize them so that all priors are comparable and lie in $(0,1)$. We first unify the direction so that ``larger means more important'' (for cosine, use $1-\cos$). Next, we apply heavy-tail compression on nonnegative metrics via $\log(1+x)$, then a robust $z$-score using the median and IQR:
\begin{equation}
 z_{\ell}=\frac{\widetilde{x}_{\ell}-\mathrm{median}_x}{\mathrm{IQR}_x+\varepsilon}.
\end{equation}
Finally, we map the robust scores to $(0,1)$ via rank normalization (empirical CDF) across layers, producing a unitless prior $\tilde{p}$. We fuse $\tilde{p}$ with the learned scores in \Cref{eq:formulation}. In our default setting, we compute priors on the WikiText-2 validation set for offline importance estimation; we report additional results with alternative estimation sets in the Appendix \ref{app:prior-normalization}.

\subsubsection{Implementation}
The Decision Module is trained jointly with the base model under the standard SFT objective. We pool token representations, score candidate layers with a two-layer MLP, convert the normalized budget into an integer $k$, and apply hard Top-$k$ selection to obtain a binary execution mask over intermediate layers. The mask gates each layer's residual update (execute if $M_\ell=1$, skip otherwise). We adopt a straight-through estimator (STE) \citep{bengio2013estimating}: the forward pass uses the hard mask, while the backward pass uses a continuous relaxation so gradients from the language-modeling loss flow into the router. More details are provided in Appendix \ref{app:decison_module}.

\subsection{KV-Aware Planner}\label{sec:kv_aware}

During autoregressive inference, an LLM generates one token at a time. If the Decision Module were to consume only the newly generated token at each decoding step, it would lack sufficient context and thus struggle to choose an effective execution path. To address this, we \emph{reuse the first layer's KV cache} to provide a lightweight global summary of the full prefix. Therefore, we concatenate the KV values of the newly generated tokens with those stored in the cache to obtain global contextual information. 
We use $G_t \in \{K_t, V_t, K_t + V_t\}$ (discussed in the ablation study in \Cref{sec:ablation}) as the input to the Decision module, enabling dynamic prediction of the optimal reasoning path during the decoding stage.

\subsection{Budget Predictor}\label{sec:budget_predictor}
When the user does not provide a budget $b$, we predict a discrete depth level from the input. Because routing is performed at the block level, we discretize budgets into an action space that specifies how many middle layers to execute:
$A=\{1,2,\dots,L_{\mathrm{mid}}\}$, where $L_{\mathrm{mid}}$ is the number of candidate depth levels. The Budget Predictor takes the same KV-aware context $G_t$ (\Cref{sec:kv_aware}) and outputs a categorical policy $\pi_\theta(k\mid G_t)$ over $A$. We choose
\begin{equation}
\hat{k}=\arg\max_{k\in A}\pi_\theta(k\mid G_t),\qquad
\hat{b}=\frac{\hat{k}+2}{L},
\end{equation}
where ``$+2$'' accounts for the always-executed first and last layers. We train the predictor with GRPO \citep{shao2024deepseekmath} using a reward that trades off predictive quality and compute; details are provided in Appendix \ref{app:grpo_training}.

%% file: pages/05_evaluation.tex
\begin{table*}[ht]
\centering
\caption{
Performance comparison on Qwen2.5-7B-Instruct across eight commonsense reasoning benchmarks for different pruning methods at various sparsity levels (14.3\%-57.1\%). \textbf{Boldface} indicates the best performance, and the \underline{Underline} means the second-order performance. \textsc{Buddy} achieves the best or competitive average performance across pruning ratios while using a single multi-budget model.
}
\label{tab:result_qwen2.5}
\resizebox{\textwidth}{!}{
\begin{tabular}{l|c|c|ccccccccc}
\toprule
Method & Pruning & RM Blocks & OBQA & PIQA & BoolQ & SIQA & Hellaswag & ARC-E & ARC-C & Winogrande & Avg. \\
\midrule
Dense w/o & - & 0(0.0\%) & 48.80 & 80.30 & 86.33 & 51.59 & 80.48 & 81.94 & 54.95 & 70.96 & 69.42 \\
\midrule
Shortened Llama & static  & 4(14.3\%) & \underline{44.80} & \textbf{80.36} & 78.75 & 48.26 & \underline{73.22} & 77.86 & 48.63 & 65.19 & 64.63 \\
ShortGPT        & static  & 4(14.3\%) & 44.00 & 77.42 & \underline{82.39} & \textbf{50.61} & 72.95 & \underline{79.25} & \underline{50.51} & \textbf{66.77} & \underline{65.49} \\
SLEB            & static  & 4(14.3\%) & \textbf{45.20} & \underline{79.11} & 76.24 & 48.57 & \textbf{73.25} & 77.90 & 50.34 & 62.90 & 64.19 \\
PuDDing         & dynamic & 4(14.3\%) & 40.60 & 75.84 & 71.01 & 45.65 & 69.30 & 72.56 & 46.67 & 59.04 & 60.08 \\
FiRST           & dynamic & 4(14.3\%) & 33.40 & 55.44 & 68.65 & 37.97 & 44.28 & 49.16 & 36.18 & 51.14 & 47.03 \\
\rowcolor{gray!12}\textsc{Buddy}           & dynamic & 4(14.3\%) & 43.40 & 77.26 & \textbf{84.50} & \underline{48.93} & 73.20 & \textbf{79.88} & \textbf{51.19} & \underline{66.69} & \textbf{65.63} \\
\midrule
Shortened Llama & static  & 8(28.6\%) & 39.80 & \underline{74.48} & \underline{65.14} & \underline{45.29} & \underline{62.44} & 67.42 & 40.70 & 57.77 & 56.63 \\
ShortGPT        & static  & 8(28.6\%) & \underline{41.00} & 73.72 & 56.94 & 43.96 & 62.43 & \underline{70.08} & \underline{41.13} & \underline{59.83} & 56.14 \\
SLEB            & static  & 8(28.6\%) & \textbf{41.60} & \textbf{75.95} & 55.47 & 43.71 & \textbf{63.86} & \textbf{74.24} & \textbf{42.58} & 56.35 & \underline{56.72} \\
PuDDing         & dynamic & 8(28.6\%) & 32.20 & 70.89 & 56.02 & 40.23 & 53.64 & 65.11 & 35.41 & 54.85 & 51.04 \\
FiRST           & dynamic & 8(28.6\%) & 26.00 & 52.34 & 55.96 & 36.18 & 35.43 & 39.52 & 28.07 & 49.01 & 40.32 \\
\rowcolor{gray!12}\textsc{Buddy}           & dynamic & 8(28.6\%) & 36.40 & 72.52 & \textbf{72.78} & \textbf{46.42} & 59.92 & 68.90 & 40.02 & \textbf{60.77} & \textbf{57.22} \\
\midrule
Shortened Llama & static  & 12(42.9\%) & 33.20 & \textbf{71.98} & 46.61 & \underline{41.91} & \textbf{51.90} & \textbf{65.36} & \underline{33.02} & 50.91 & 49.36 \\
ShortGPT        & static  & 12(42.9\%) & \textbf{37.00} & 67.36 & \underline{57.68} & 41.56 & 48.00 & 59.26 & 32.42 & 52.41 & 49.46 \\
SLEB            & static  & 12(42.9\%) & \underline{35.60} & \underline{70.35} & 51.16 & \underline{41.91} & \underline{51.23} & \underline{64.56} & \underline{33.02} & \underline{52.88} & \underline{50.09} \\
PuDDing         & dynamic & 12(42.9\%) & 32.00 & 60.77 & 50.80 & 36.54 & 36.54 & 46.72 & 26.19 & 48.78 & 42.29 \\
FiRST           & dynamic & 12(42.9\%) & 31.80 & 52.77 & 57.61 & 36.13 & 38.32 & 41.08 & 29.52 & 50.12 & 42.17 \\
\rowcolor{gray!12}\textsc{Buddy}           & dynamic & 12(42.9\%) & 33.20 & 67.03 & \textbf{60.70} & \textbf{41.97} & 49.25 & 60.27 & \textbf{33.36} & \textbf{56.59} & \textbf{50.30} \\
\midrule
Shortened Llama & static  & 16(57.1\%) & 29.20 & 63.06 & 57.49 & 37.10 & 37.34 & 49.16 & 26.11 & 50.91 & 43.80 \\
ShortGPT        & static  & 16(57.1\%) & 27.20 & 57.13 & \underline{60.76} & 36.95 & 32.13 & 36.24 & 24.91 & 49.41 & 40.59 \\
SLEB            & static  & 16(57.1\%) & \underline{30.60} & \textbf{66.49} & 45.38 & \textbf{40.23} & \textbf{40.50} & \textbf{56.02} & \underline{27.39} & \textbf{53.12} & \underline{44.96} \\
PuDDing         & dynamic & 16(57.1\%) & 27.80 & 57.13 & 51.07 & 34.85 & 30.35 & 34.85 & 23.63 & 51.93 & 38.95 \\
FiRST           & dynamic & 16(57.1\%) & 27.00 & 51.41 & \textbf{62.17} & 34.24 & 26.67 & 25.29 & 24.57 & 50.20 & 37.69 \\
\rowcolor{gray!12}\textsc{Buddy}           & dynamic & 16(57.1\%) & \textbf{31.80} & \underline{63.98} & 59.08 & \underline{37.21} & \underline{38.15} & \underline{50.38} & \textbf{27.99} & \underline{52.33} & \textbf{45.11} \\
\bottomrule
\end{tabular}}
\end{table*}

\section{Experiments}

\subsection{Settings}\label{experiment_settings}

\paragraph{LLMs.} 
We evaluate \textsc{Buddy} on Llama3-8B,\citep{touvron2023llama}, and Qwen2.5-7B \citep{yang2024qwen} (see Appendix \ref{appendix:LLM} for exact versions).For 32-layer Llama-family models, we evaluate removing 4/8/12/16 blocks, corresponding to 12.5\%, 25\%, 37.5\%, and 50\% depth sparsity. For Qwen2.5-7B-Instruct, which has 28 layers, the same removal counts correspond to 14.3\%, 28.6\%, 42.9\%, and 57.1\% sparsity.

\paragraph{Benchmark.}
We conduct experiments for these LLMs on the \textbf{Commonsense Reasoning} benchmarks, including BoolQ \citep{clark2019boolq}, PIQA \citep{bisk2020piqa}, HellaSwag \citep{zellers2019hellaswag}, WinoGrande \citep{sakaguchi2021winogrande}, ARC-easy and ARC-challenge \citep{clark2018think}, OpenbookQA \citep{mihaylov2018can}, and SIQA \citep{sap2019siqa}. We employed lm-eval-harness \citep{eval-harness} to create open prompts for the benchmarks and produce the results.

\paragraph{Baselines.}
We compare against recent depth-pruning approaches. \emph{Static} methods:
\textbf{(1) Shortened LLaMA} \citep{kim2024shortened}, rank blocks $\Delta\text{PPL}$ ;
\textbf{(2) ShortGPT} \citep{men2024shortgpt}, rank by input–output cosine similarity ;
\textbf{(3) SLEB} \citep{song2024sleb}, iterative rank layers by $\Delta\text{PPL}$. 
\emph{Dynamic} methods:
\textbf{(4) PuDDing} \citep{wee2025pudding}, construct the omission set from the evaluated commonsense benchmarks; 
\textbf{(5) FiRST} \citep{jain2025first}, add lightweight per-layer linear routers to predict execute/skip decisions. For details on reproduction, please refer to Appendix \ref{app:baselines}.

\paragraph{\textbf{Hyper-parameters and Training Details.}}
We adopted the popular LoRA ($r=\text{8}$, all linear modules) method to fine-tune the LLMs. We used Alpaca \citep{taori2023stanford} dataset to fine-tune the models. We applied the Value States as the features and adopted the Taylor metric as the prior-knowledge, and set the $\omega$ factor to 0.1. We converted the model precision to BFloat16 and used AdamW as the optimizer with 100 warm-up steps and trained the model with a learning rate of \(1 \times 10^{-4}\) and batch size 8 for 2 epochs. We apply the same training configuration across each baseline. Full training details are in Appendix \ref{app:implementation}.

\begin{table*}[t]
\centering
\caption{Speed analysis on the Alpaca and SAMSum datasets. The speed is measured as tokens/s.}
\label{tab:result_speed}
\resizebox{\textwidth}{!}{
\begin{tabular}{l|c|cccc|cccc}
\toprule
 \multirow{2}{*}{Method} & \multirow{2}{*}{RM Blocks} & \multicolumn{4}{c|}{Alpaca} & \multicolumn{4}{c}{SAMSum} \\
\cmidrule(r){3-10}
 & & Prefill & Speed up & Decode & Speed Up & Prefill & Speed Up & Decode & Speed Up \\
\midrule
Dense & 0(0.0\%) & 1753.16 & $\times$1.00 & 39.48 & $\times$1.00 & 4033.83 & $\times$1.00 & 60.70 & $\times$1.00 \\
\midrule
\textsc{Buddy} & 4(12.5\%) & 1991.54 & $\times$1.14 & 40.03 & $\times$1.01 & 4035.68 & $\times$1.00 & 61.28 & $\times$1.01 \\
\textsc{Buddy} & 8(25\%) & 2440.14 & $\times$1.39 & 46.80 & $\times$1.19 & 4777.07 & $\times$1.18 & 71.91 & $\times$1.18 \\
\textsc{Buddy} & 12(37.5\%) & 2548.39 & $\times$1.45 & 51.49 & $\times$1.30 & 5476.04 & $\times$1.36 & 79.11 & $\times$1.30 \\
\textsc{Buddy} & 16(50\%) & 3348.18 & $\times$1.91 & 64.84 & $\times$1.64 & 6581.55 & $\times$1.63 & 99.12 & $\times$1.63 \\
\bottomrule
\end{tabular}}
\end{table*}

\paragraph{Main Results.}
Across eight commonsense reasoning benchmarks and four sparsity settings, \textbf{\textsc{Buddy}} achieves strong average accuracy and degrades gracefully as pruning increases. It is the best-performing method in most settings and remains competitive in the remaining cases, while using a single trained model to serve multiple budgets. (\Cref{tab:result_llama8b}--\Cref{tab:result_qwen2.5}). On Llama3-8B, \textbf{\textsc{Buddy}} retains approximately 99.9\%, 90.8\%, 80.7\%, and 71.3\% of original accuracy at 12.5\%, 25\%, 37.5\%, and 50\% sparsity, respectively, outperforming both static and dynamic baselines at matched sparsity. On Qwen2.5-7B-Instruct, performance drops are larger overall, but \textbf{\textsc{Buddy}} remains among the best methods across sparsity levels. Results on additional backbones (Llama2-7B and Llama1-13B) are reported in Appendix \ref{app:more_results}. Overall, \textbf{\textsc{Buddy}} attains the best average accuracy in most settings while supporting multiple budgets with a single trained model.


\begin{figure*}[!b]
\centering

\begin{minipage}{0.33\linewidth}
\centering
\includegraphics[width=\linewidth]{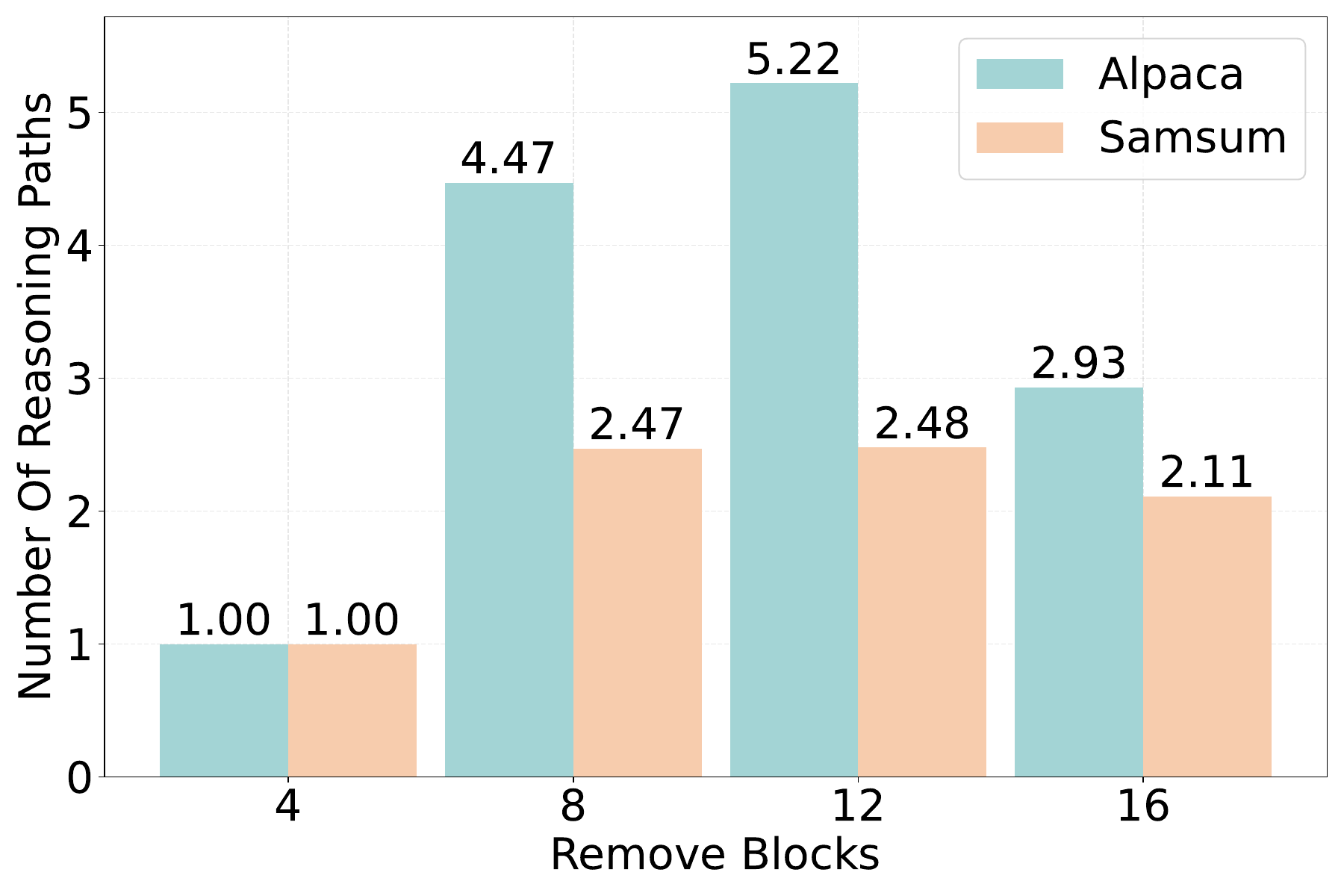}
\caption{Inference path count.}
\label{fig:sub:decode_decision_bar}
\end{minipage}
\hfill
\begin{minipage}{0.66\linewidth}
\centering
\includegraphics[width=\linewidth]{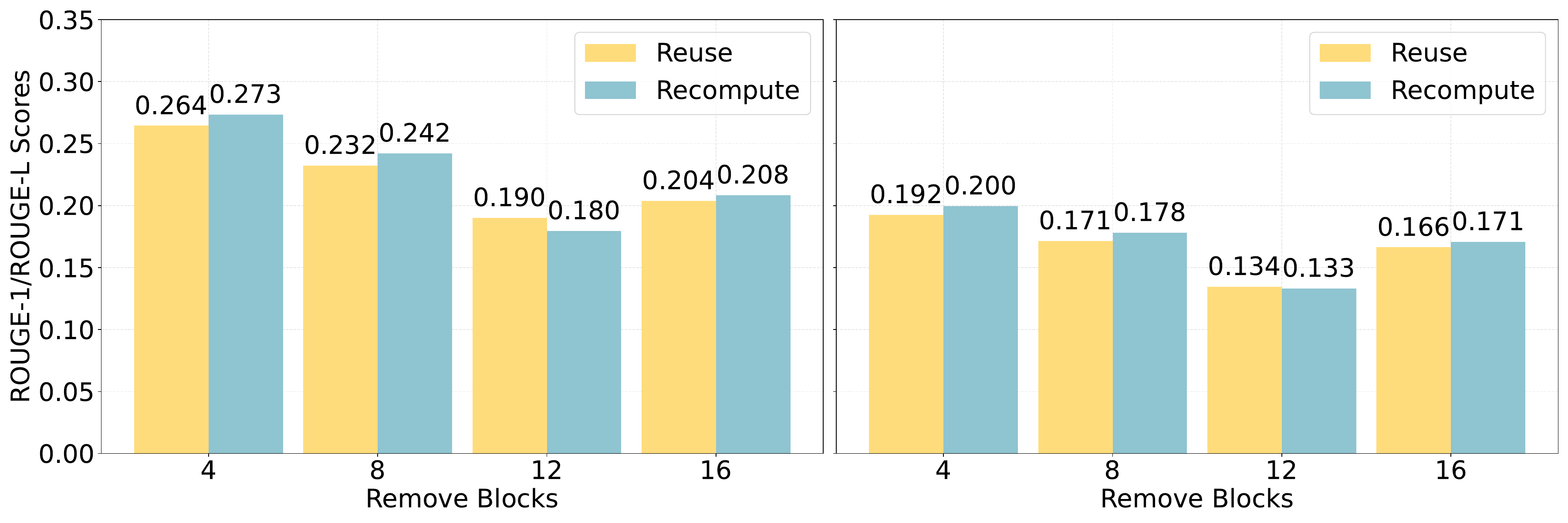}
\caption{ROUGE-1 (left) and ROUGE-L (right) scores between reuse and recomputation.}
\label{fig:sub:decode_acc_bar}
\end{minipage}

\end{figure*}


\subsection{Analysis}

\subsubsection{Speed Analysis}
We measure end-to-end throughput (tokens/s) on Alpaca and SAMSum for both \emph{prefill} and \emph{decode}. During decoding, we generate 128 tokens. Table \ref{tab:result_speed} shows: (i) speedups are consistently larger in prefill than in decode; (ii) Alpaca exhibits larger prefill gains than SAMSum at matched sparsity; and (iii) speedups increase with sparsity. At very light pruning (12.5\%), per-step routing and gather/scatter overhead offsets (analysized on Appendix \ref{app:inferenece_overhead}) most savings (especially on SAMSum), whereas beyond $\sim$25\% sparsity the benefits clearly outweigh overhead.

\subsubsection{Layer Selection Analysis}
\begin{figure*}[!t]
\centering
\includegraphics[width=\linewidth]{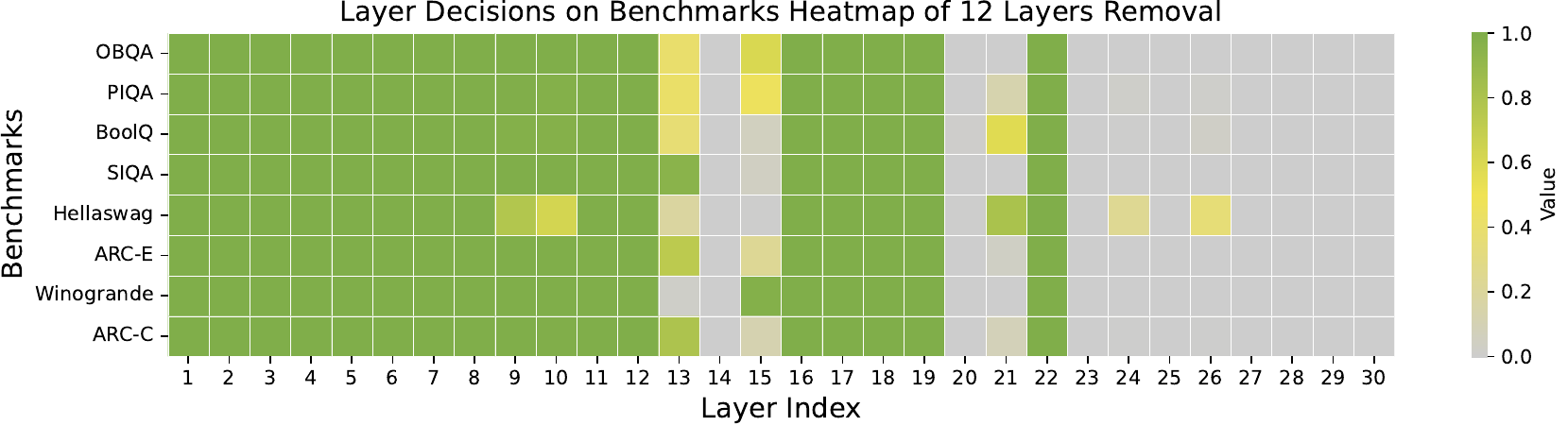}
\caption{
A visual illustration of the \textsc{Buddy}'s average decision of each transformer block on 37.5\% sparsity (12 layers removed). The color \textcolor{darkgreen}{green} indicates that the blocks are likely important to be maintained, the color \textcolor{darkgrey}{grey} indicates that the blocks are likely to be pruned, and the color \textcolor{darkyellow}{yellow} means the importance of layers varies across tasks and samples. 
}
\label{fig:heat_map_rm12}
\end{figure*}

We analyze routing decisions across tasks and inputs. \Cref{fig:heat_map_rm12} summarizes, at 37.5\% sparsity, the per-layer execution frequency averaged over inputs from eight benchmarks. Several blocks are consistently selected (e.g., layers 1--8 and 16--17), while others are frequently skipped (e.g., layers 14, 23, 25, and 27--30). The remaining blocks vary by task: for example, layers 13, 15, and 21 exhibit large variability, being crucial for some datasets but dispensable for others. These results support input- and task-adaptive routing. More results are shown in Appendix \ref{app:speed_analysis}.

\subsubsection{Decode Adaptive Analysis}
\paragraph{Inference path counts.}
We examine how often the execution path changes during decoding. For Alpaca and SAMSum, we run inference at multiple sparsity levels and count the number of distinct layer masks observed across decoding steps. As shown in \Cref{fig:sub:decode_decision_bar}, path changes are rare at low sparsity (remove 4 blocks), indicating that light pruning yields a largely stable route. At moderate sparsity, the number of distinct paths increases substantially, reflecting stronger context-driven adaptation. The effect is more pronounced on Alpaca than on SAMSum, consistent with longer prompts and richer context evolution in Alpaca.

\paragraph{Performance comparison.}
We explore whether path recomputation during decoding helps accuracy. We compare two strategies on SAMSum using ROUGE \citep{lin2004rouge}: \emph{Reuse}, which fixes the decode path to the prefill decision, versus \emph{Recompute}, which updates the path at each step using fresh hidden states. As in \Cref{fig:sub:decode_acc_bar}, \emph{Recompute} consistently outperforms \emph{Reuse} on ROUGE-1 and ROUGE-L, indicating that adapting the route to the evolving context yields better summaries without increasing the executed-layer budget.

\subsection{Ablation Study}\label{sec:ablation}
\paragraph{\textbf{Hidden-State Features.}} 
We compare three router inputs extracted from the first-layer KV cache: (1) Key states, (2) Value states, and (3) Key+Value via element-wise sum. \Cref{tab:ablation_states} shows that Value states yield the best accuracy at light-to-moderate pruning (12.5\%--37.5\%), while at extreme sparsity (50\%) Key states perform slightly better. The Key+Value variant underperforms, suggesting redundancy and more difficult calibration for the scorer.

\begin{table}[htbp]
\centering
\caption{
Ablation study on Decision Module input using different hidden state representations. Average accuracy (\%) is reported.
}
\label{tab:ablation_states}
\resizebox{\linewidth}{!}{
\begin{tabular}[t]{c|ccc}
  \toprule
  RM Blocks & Key States & Value States & Key+Value States \\ 
  \midrule
  4(12.5\%)  & \underline{62.12} & \textbf{62.21} & 60.92 \\
  8(25.0\%)  & \underline{55.71} & \textbf{58.32} & 55.27 \\
  12(37.5\%) & \underline{50.10} & \textbf{52.44} & 49.34 \\
  16(50.0\%) & \textbf{45.15} & \underline{44.78} & 43.67 \\
  \bottomrule
\end{tabular}}
\end{table}

\paragraph{\textbf{Prior Knowledge Fusion.}} 
We ablate the layer-importance priors used in score fusion: (1) None, (2) $\Delta\text{PPL}$, (3) Cosine dissimilarity, and (4) Taylor/Fisher. Results in \Cref{tab:ablation_prior} show \textbf{Taylor} consistently performs best across pruning ratios. $\Delta\text{PPL}$ lags at mild pruning but becomes competitive at high sparsity (50\%). Overall, strong priors help, especially when compute is tight.

\begin{table}[htbp]
\centering
\caption{
Ablation study on prior knowledge integration for score fusion in the Decision Module. Average accuracy (\%) is reported.
}
\label{tab:ablation_prior}
\resizebox{\linewidth}{!}{
\begin{tabular}[t]{c|cccc}
  \toprule
  RM Blocks & None & $\Delta\text{PPL}$ & Cosine Similarity & Taylor \\ 
  \midrule
  4(12.5\%)  & \underline{62.21} & 60.54 & 62.20 & \textbf{62.63} \\
  8(25.0\%)  & \underline{58.32} & 55.03 & 58.27 & \textbf{58.44} \\
  12(37.5\%) & 52.44 & 49.98 & \underline{53.27} & \textbf{53.54} \\
  16(50.0\%) & 44.78 & \underline{46.45} & 45.93 & \textbf{46.90} \\
  \bottomrule
\end{tabular}}
\end{table}

\paragraph{\textbf{Omega Coefficient.}} 
We conducted ablation experiments on the $\omega$ hyperparameter in the Decision Module under different pruning ratios, testing six different $\omega$ values: 0.01, 0.1, 0.3, 0.5, 0.7, and 1.0. As shown in \Cref{tab:ablation_lambda}, $\omega$=0.1 achieved the best performance across all pruning settings. In addition, we found that as $\omega$ increases, performance gradually decreases, meaning that prior knowledge fusion only requires a small amount of guidance, and excessive participation will lead to performance degradation.

\paragraph{\textbf{Layer Selection for Feature Extraction.}} 
We conduct an ablation study to investigate how the placement of the Decision Module, i.e., extracting contextual information from different transformer layers, affects routing quality. Specifically, we use the KV cache from layers 1 to 4 as the context input on Llama2-7B, and compare the average performance on the commonsense benchmark, as reported in \Cref{tab:ablation_start_layer}. From the results, we observe that at low sparsity levels (12.5\% and 25\%), using the first-layer KV cache achieves the best performance, which validates our original design choice. At higher sparsity levels, the third-layer KV cache yields slightly better results. Overall, choosing the first-layer KV cache as the context-extraction layer is a strong and robust default. When more layers are pruned, however, leveraging a slightly deeper layer as the context-extraction layer can provide richer contextual representations for routing decisions, and thus further improve performance under high sparsity.

\begin{table}[htb]
\centering
\caption{
Ablation on the balance coefficient $\omega$. Average accuracy (\%) is reported.
}
\label{tab:ablation_lambda}
\resizebox{\linewidth}{!}{
\begin{tabular}[t]{c|cccccc}
  \toprule
  RM Blocks & $\omega$=0.01 & $\omega$=0.1 & $\omega$=0.3 & $\omega$=0.5 & $\omega$=0.7 & $\omega$=1.0 \\ 
  \midrule
  4 (12.5\%)  & \underline{62.24} & \textbf{62.63} & 62.09 & 62.07 & 61.84 & 61.57 \\
  8 (25.0\%)  & \underline{57.94} & \textbf{58.44} & 57.90 & 57.89 & 57.68 & 56.99 \\
  12 (37.5\%) & 48.83 & \textbf{53.54} & 52.17 & 51.38 & 44.48 & \underline{53.18} \\
  16 (50.0\%) & 42.52 & \textbf{46.90} & \underline{43.03} & 34.85 & 35.30 & 34.61 \\
  \bottomrule
\end{tabular}}
\end{table}

%% file: pages/06_conclusion.tex
\vspace{-10pt}
\section{Conclusion}

In this paper, we introduce \textsc{Buddy}, a budget-driven, decode-adaptive depth routing framework for LLM inference. A Decision Module automatically selects the most suitable layers for inference based on user input and budget constraints. A KV-aware planner reuses the KV cache to update the model's inference path during the autoregressive inference stage. A Budget Predictor automatically determines the optimal budget based on user input when no explicit budget is provided. The framework updates routes during decoding and honors explicit or predicted budgets. Overall, \textsc{Buddy} targets an accuracy–efficiency–flexibility trade-off for heterogeneous-budget LLM serving. Since a single trained model can adapt to different budget requirements, \textsc{Buddy} reduces the need to train separate sparsity-specific models and avoids deploying multiple checkpoints simultaneously, lowering both training cost and system-level GPU memory overhead.

\paragraph{\textbf{Limitations and future work}}
Because our method updates the execution path during inference to preserve performance, \emph{all} Transformer blocks remain resident in GPU memory; thus, VRAM usage does not decrease. Besides, for batched inputs, different sequences may select different paths, which can induce KV-cache misses at skipped layers. In this paper, we use zero-filled vectors to indicate ``non-executed'' states; in future work we plan to develop memory-aware routing and more efficient KV-cache strategies (e.g., selective cache sharing/compaction, lightweight block swapping, and batch-level path grouping) to better support adaptive inference at scale.

\section*{Impact Statement}
This paper presents work whose goal is to advance the field of machine learning by enhancing the efficiency and budget-flexibility of LLM inference. By reducing the computational resources required for model execution, this research contributes to the environmental sustainability of AI and potentially lowers the barrier for accessing advanced language technologies. We do not foresee any specific negative societal or ethical consequences that distinguish this work from general advancements in model compression and adaptive inference.

%% file: pages/appendix.tex
\appendix
\section*{Appendix}

\section{Observations}

\begin{figure}[!h]
\centering
\includegraphics[width=0.8\linewidth]{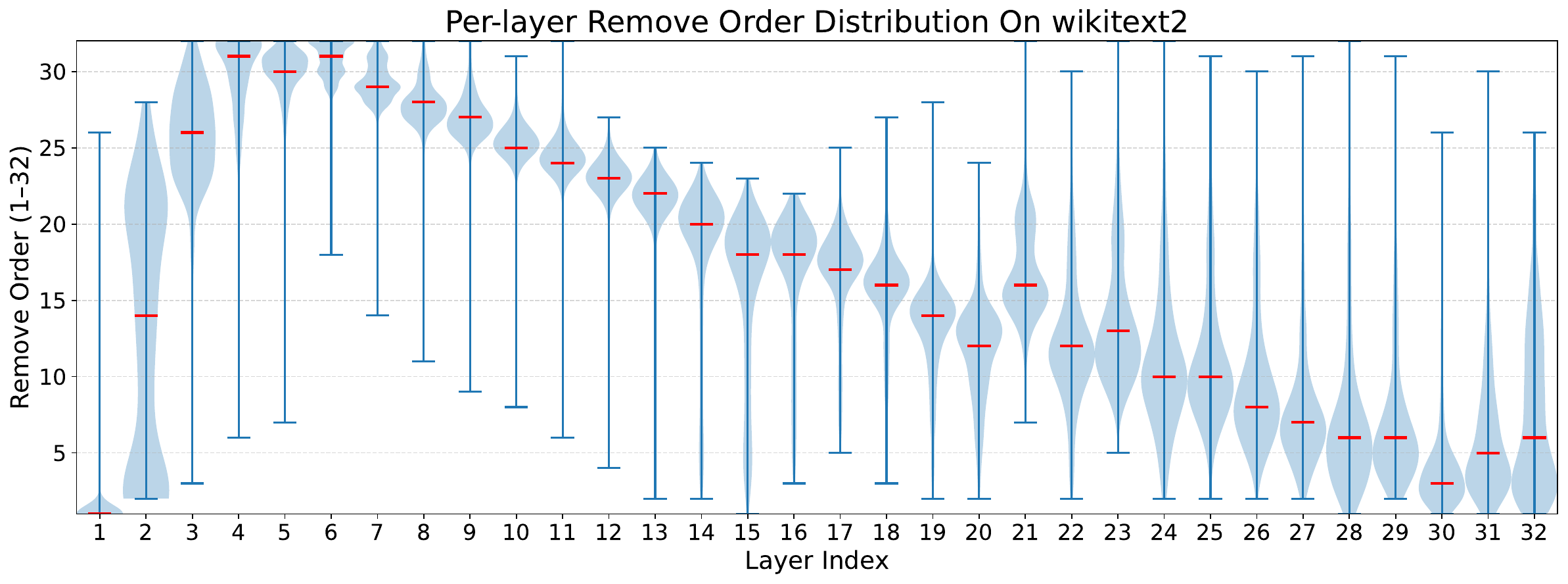}
\caption{
Input-dependent layer importance ranking distributions across different inputs on the WikiText-2 datasets. The chart shows how layer rankings vary significantly across inputs, demonstrating the necessity for dynamic pruning decisions. 
}
\label{fig:observation_1}
\end{figure}

We also adopt Taylor score and $\Delta\mathrm{PPL}$ as the evaluation metric on the Llama2-7B model, and compute the layer-importance distributions on the WikiText-2 and PTB datasets. The results are shown in \Cref{fig:observation_1_2} and \Cref{fig:observation_1}. They reveal that the estimated importance of different layers varies dramatically when different metrics are used (Taylor score vs. $\Delta\mathrm{PPL}$). Moreover, the importance of each layer also differs across evaluation datasets (WikiText-2 vs. PTB). These phenomena further corroborate Observation \ref{obs:observation_1}: as the input and evaluation metric change, the importance of model layers also shifts, which calls for a dynamic layer-pruning scheme that can adapt to different inputs.

\begin{figure}[!h]
\centering
\includegraphics[width=0.8\linewidth]{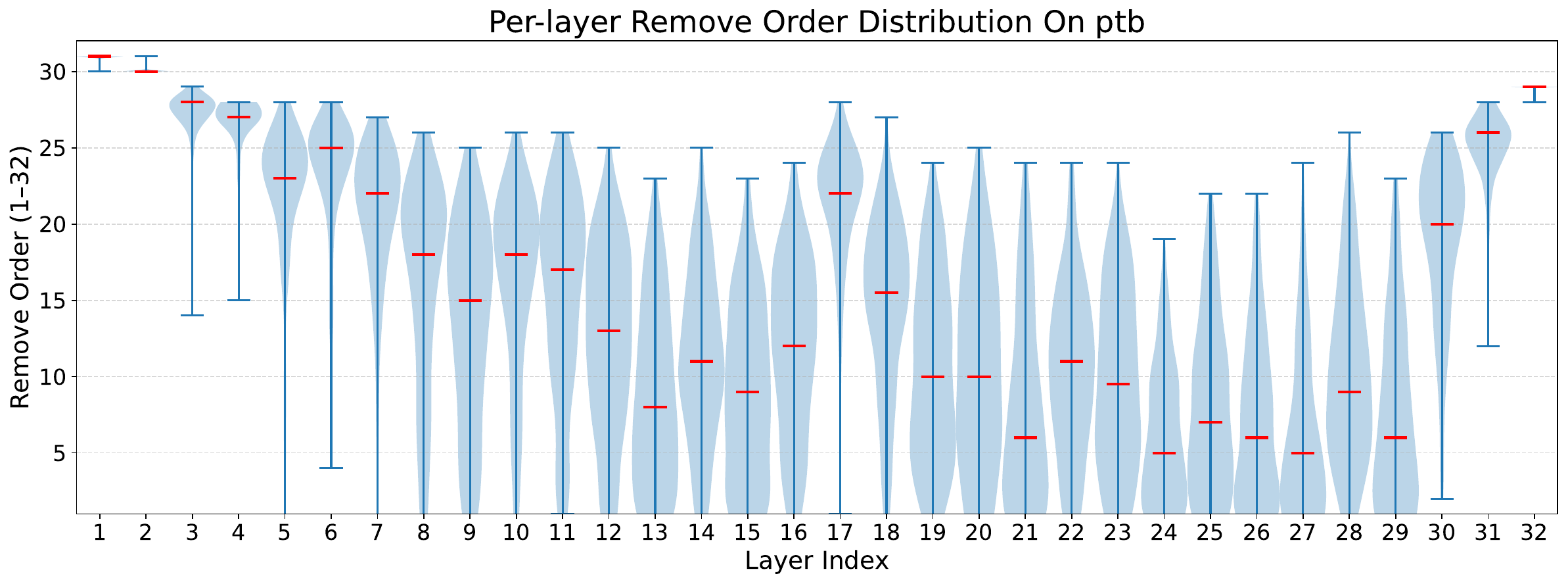}
\caption{
Input-dependent layer importance ranking distributions across different inputs on the PTB datasets. The chart shows how layer rankings vary significantly across inputs, demonstrating the necessity for dynamic pruning decisions. 
}
\label{fig:observation_1_3}
\end{figure}

\section{Prior Indicators and Normalization}\label{app:prior-normalization}
We consider $L$ transformer layers and $M$ prior indicators that estimate per-layer importance.
Let $x_{l,m}$ denote the raw score of indicator $m\in\{1,\dots,M\}$ on layer $l\in\{1,\dots,L\}$.
Different indicators can have heterogeneous scales and tails (e.g., perplexity gaps versus gradient norms), so we map them to a common, robust scale before fusion. Our pipeline has four steps, and each step is applied per indicator $m$ across layers $l$.

\paragraph{(i) Direction unification.}
Ensure ``larger $\Rightarrow$ more important'' for every indicator:
\begin{itemize}
\item \textbf{$\Delta$PPL}: $x_{l,\text{ppl}} := \max\{0,\, \mathrm{PPL}_{\text{skip}(l)}-\mathrm{PPL}_{\text{full}}\}$.
\item \textbf{Taylor/Fisher/GradNorm}: already positively oriented; clip to nonnegative.
\item \textbf{Cosine similarity}: convert to \emph{dissimilarity}
$d_{l,\cos}:=1-\cos(\cdot,\cdot)\in[0,2]$ and set $x_{l,\cos}:=d_{l,\cos}$.
\end{itemize}

\paragraph{(ii) Heavy-tail stabilization.}
For heavy-tailed indicators (typically $\Delta$PPL and Taylor/Fisher/GradNorm),
apply a monotone variance–stabilizing transform, e.g.,
\begin{equation}
\tilde{x}_{l,m} \;=\; \log\!\bigl(1 + x_{l,m}\bigr)
\quad\text{(or Box--Cox: }\tilde{x}_{l,m} = \tfrac{x_{l,m}^{\lambda}-1}{\lambda},~\lambda\in[0.2,0.5]\text{)}.
\end{equation}
For cosine dissimilarity, we typically use the identity: $\tilde{x}_{l,\cos} := x_{l,\cos}$.

\paragraph{(iii) Robust standardization and outlier control.}
Compute a robust $z$-score using median and interquartile range (IQR) across layers:
\begin{equation}
z_{l,m} \;=\; \frac{\tilde{x}_{l,m} - \mathrm{median}_l(\tilde{x}_{l,m})}
{\mathrm{IQR}_l(\tilde{x}_{l,m}) + \varepsilon},
\qquad \mathrm{IQR} := Q_{75}-Q_{25},~\varepsilon>0.
\end{equation}
Optionally winsorize: $z_{l,m}\leftarrow \mathrm{clip}(z_{l,m},-c,c)$ with $c\in[2.5,3.5]$.

\paragraph{(iv) Unit-interval mapping.}
Map $z_{l,m}$ to a comparable, dimensionless score $u_{l,m}\in(0,1)$.
We use rank normalization (scale-free and robust):
\begin{equation}
u_{l,m} \;=\; \frac{1+\mathrm{rank}_l(z_{l,m})}{L+1}
\quad\in(0,1),
\end{equation}
or, alternatively, a temperature-controlled sigmoid
$u_{l,m}=\sigma\!\bigl(z_{l,m}/\tau\bigr)$ with $\tau\in[1,2]$.

\vspace{0.25em}
\noindent
The above yields $M$ normalized importance profiles $\{u_{\cdot,m}\}_{m=1}^M$.

\subsection{Indicator-specific recipes}
\label{sec:indicator-recipes}
\paragraph{A. $\Delta$PPL (skip-induced perplexity increase).}
\begin{align}
\Delta\mathrm{PPL}_l &= \max\{0,\,\mathrm{PPL}_{\text{skip}(l)}-\mathrm{PPL}_{\text{full}}\}, \\
u_{l,\text{ppl}} &= \mathrm{RankNorm}\!\Bigl(\mathrm{RobustZ}\bigl(\log(1+\Delta\mathrm{PPL}_l)\bigr)\Bigr).
\end{align}
This emphasizes layers whose removal most degrades language modeling quality while suppressing magnitude outliers.

\paragraph{B. Taylor/Fisher/gradient-norm importance.}
\begin{align}
u_{l,\text{taylor}} &= \mathrm{RankNorm}\!\Bigl(\mathrm{RobustZ}\bigl(\log(1+\mathrm{Score}^{\text{taylor}}_l)\bigr)\Bigr).
\end{align}
Any nonnegative saliency-like score (e.g., first-order Taylor, Fisher diagonal, $\ell_2$ gradient norm) fits this template.

\paragraph{C. Cosine-similarity based discrepancy.}
Let $s^{\text{full}}_l$ and $s^{\text{skip}(l)}_l$ be per-layer summary representations (or outputs) from the full and ``skip-$l$'' forward passes, respectively. Define the cosine dissimilarity
$d_l = 1 - \cos\bigl(s^{\text{skip}(l)}_l,\, s^{\text{full}}_l\bigr)$ and set
\begin{align}
u_{l,\cos} &= \mathrm{RankNorm}\!\Bigl(\mathrm{RobustZ}(d_l)\Bigr).
\end{align}
This treats larger representational deviation as higher importance.

\section{Implementation}\label{app:implementation}

\subsection{Decision Module Design}\label{app:decison_module}
We implement the Decision Module as a lightweight two-layer MLP (Linear--SiLU--Linear). For each inference step, the module produces a real-valued score for every intermediate Transformer block. We optionally normalize the scores with a softmax to obtain a distribution over candidate blocks, and deterministically activate the top-$k$ blocks to form a binary execution mask $M\in\{0,1\}^{L}$ over layers (excluding the always-executed first and last blocks). The resulting mask is used to gate computation via the layer-skipping mechanism described in \Cref{sec:layer_skipping} yielding dynamic-depth inference without altering tensor shapes. Batched inference introduces heterogeneous execution paths across requests. To efficiently support this setting, we process each block on a per-layer sub-batch: at block $\ell$, we gather only the requests with $M_\ell=1$ and run the block on this active subset, while skipping the block for the inactive requests. This grouping-by-mask strategy preserves correctness while reducing wasted computation under mixed routing decisions within the same batch.

\subsection{Inference overhead}\label{app:inferenece_overhead}

We first analyze the theoretical latency of the Decision Module. We denote by $d$ the hidden size, by $d_c$ the MLP intermediate size (for Llama-7B, $d_c=11008$ when $d=4096$), by $T$ the input length, by $L$ the total number of transformer layers, by $I$ the number of ignored (fixed) layers in the router, by $r$ the hidden width of the Decision Module MLP, by $B$ the batch size, and by $b\in(0,1]$ the budget parameter that controls the effective fraction of active layers. The Decision Module takes an input $x\in\mathbb{R}^{B\times T\times d}$ and applies two linear layers token-wise, $d\to r$ and $r\to(L-I)$, followed by a softmax and Top-$k$ whose FLOPs are negligible compared to the dense matrix multiplications. Counting one multiply–add as $2$ FLOPs, its cost is approximated by \[ F_{\mathrm{DM}} \approx 2BTdr + 2BTr(L-I). \] For the backbone transformer, let $F_{\text{layer}}$ denote the FLOPs of a single decoder layer with hidden size $d$ and sequence length $T$. A Llama-style decoder layer has: three projections $Q,K,V$ with cost $3\times 2Td^2 = 6Td^2$, one output projection with cost $2Td^2$, attention score/product $QK^\top$ and $AV$ with cost approximately $4T^2d$, and an MLP with architecture $d\to d_c\to d$, whose two linear layers cost $2Td d_c + 2T d_c d = 4T d d_c$. Thus the corrected per-layer FLOPs with $d\to d_c\to d$ are \[ F_{\text{layer}} \approx (6Td^2 + 2Td^2) + 4T^2d + 4Td d_c = 8Td^2 + 4T^2d + 4Td d_c. \] Executing all $L$ layers (no skipping) costs \[ F_{\text{full}} \approx L F_{\text{layer}} = L\bigl(8Td^2 + 4T^2d + 4Td d_c\bigr). \] 

The router converts the budget $b$ into a Top-$k$ via $k(b) = bL - I$ (clamped to $[1, L-I]$), so that roughly $bL$ out of the $L$ layers are effectively active (the $I$ ignored layers are always run and about $k(b)$ of the remaining $L-I$ layers are selected). Accordingly, the backbone FLOPs under budget $b$ are approximated as \[ F_{\text{model}}(b) \approx b F_{\text{full}} = bL\bigl(8Td^2 + 4T^2d + 4Td d_c\bigr). \] The relative FLOPs overhead of the Decision Module at budget $b$ is then \[ \rho(b) \approx \frac{F_{\text{DM}}}{F_{\text{DM}} + F_{\text{model}}(b)} \approx \frac{F_{\text{DM}}}{bL\bigl(8Td^2 + 4T^2d + 4Td d_c\bigr)}, \] where the second approximation uses that $F_{\text{DM}}\ll F_{\text{model}}(b)$ in practice. 

Instantiating with the configuration $L = 32$, $d = 4096$, $d_c = 11008$ (Llama-7B MLP), $T = 1024$, $r = 32$, $I = 2$, $B = 1$, we obtain for the Decision Module $F_{\text{DM}} \approx 2\cdot 1\cdot 1024\cdot 4096\cdot 32 + 2\cdot 1\cdot 1024\cdot 32\cdot(32-2) = 270{,}401{,}536 \approx 2.70\times 10^8.$ For a single transformer layer the corrected cost is $ F_{\text{layer}} \approx 8\cdot 1024\cdot 4096^2 + 4\cdot 1024^2\cdot 4096 + 4\cdot 1024\cdot 4096\cdot 11008 \approx 339{,}302{,}416{,}384 \approx 3.39\times 10^{11}$. 
Under budget $b=0.875$, the backbone FLOPs are \[ F_{\text{model}}(0.875) \approx 0.875\cdot F_{\text{full}} \approx 9{,}500{,}467{,}658{,}752 \approx 9.50\times 10^{12}, \] giving a Decision Module FLOPs ratio of \[ \rho(0.875) \approx \frac{F_{\text{DM}}}{F_{\text{model}}(0.875)} \approx \frac{2.70\times 10^8}{9.50\times 10^{12}} \approx 2.85\times 10^{-5} \approx 0.00285\%. \] 
Under budget $b=0.5$, the backbone FLOPs become \[ F_{\text{model}}(0.5) \approx 0.5\cdot F_{\text{full}} \approx 5{,}428{,}838{,}662{,}144 \approx 5.43\times 10^{12}, \] leading to \[ \rho(0.5) \approx \frac{F_{\text{DM}}}{F_{\text{model}}(0.5)} \approx \frac{2.70\times 10^8}{5.43\times 10^{12}} \approx 4.98\times 10^{-5} \approx 0.00498\%. \] Therefore, the Decision Module’s routing overhead is negligible compared to the main transformer.

\begin{table}[h]
\centering
\caption{Latency breakdown of \textsc{Buddy} on an NVIDIA A100 GPU with Llama-2-7B, batch size $=5$, and sequence length $=256$.}
\label{tab:latency_breakdown}
\begin{tabular}{lcc}
\toprule
Component & Latency (ms) & Percentage (\%) \\
\midrule
Backbone Compute & 64.11 & 79.50 \\
Decision Module & 4.58 & 5.67 \\
Gather/Scatter Overhead & 2.64 & 3.28 \\
\bottomrule
\end{tabular}
\end{table}

We further analyze the runtime overhead of \textsc{Buddy}, as summarized in \Cref{tab:latency_breakdown}. On an NVIDIA A100 GPU with Llama-2-7B (\textit{batch size} $=5$, \textit{sequence length} $=256$), the backbone computation dominates the overall latency, accounting for $64.11$ ms ($79.50\%$) of the total runtime. In contrast, the decision module introduces only a small overhead of $4.58$ ms ($5.67\%$), and the gather/scatter operations required by grouping-by-mask add another $2.64$ ms ($3.28\%$). These results show that the routing-related system overhead of \textsc{Buddy} is modest in practice.


\subsection{Training Details and Overhead }
We train the Decision Module using standard supervised fine-tuning (SFT) jointly with the backbone model and LoRA adapters. Concretely, we fine-tune on the Alpaca instruction-following dataset using the Adam optimizer, and minimize the next-token cross-entropy loss. During this process, the parameters of the Decision Module and the LoRA parameters of the backbone are updated simultaneously, following the same optimization hyperparameters as in the main LoRA fine-tuning setup described in \Cref{experiment_settings}. To make the discrete layer-selection decisions trainable, we adopt a straight-through estimator (STE). In the forward pass, the Decision Module produces per-layer scores, and we apply a hard Top-$k$ gating to obtain binary execute/skip decisions. In the backward pass, we use the continuous scores to approximate the gradient, allowing gradients from the language modeling loss to flow through the gating mechanism and update the MLP parameters.

In order to enable the model to both strictly and flexibly adapt to different compute budgets, we perform \emph{random budget sampling} during training. For each mini-batch, we sample target sparsity levels from the same range used at inference time, and assign (potentially different) budgets to individual examples within the batch. This encourages the Decision Module to learn routing policies that are robust across a variety of budgets instead of overfitting to a single sparsity setting. We also explicitly include the zero-sparsity case (i.e., all layers executed) in the sampling space, so that every Transformer block is regularly activated during training and receives gradient updates, which improves overall training stability and robustness.



\begin{table}[ht]
\centering
\small
\caption{Training cost of \textsc{Buddy} across different backbone models. The training pipeline consists of two stages: (1) joint fine-tuning of the decision module and LoRA parameters, and (2) a separate GRPO stage for the budget predictor.}
\label{tab:training_cost}
\begin{tabular}{lcc}
\toprule
Model & Fine-tuning Time (h) & GRPO Time (h) \\
\midrule
Llama2-7B   & 10.46 & 0.90 \\
Llama3-8B   & 8.26  & 0.82 \\
Llama1-13B  & 24.93 & 1.49 \\
Qwen2.5-7B  & 11.21 & 0.78 \\
\bottomrule
\end{tabular}
\end{table}

We report the end-to-end training cost of \textsc{Buddy} in \Cref{tab:training_cost}. The training procedure consists of two stages: (1) joint fine-tuning of the decision module and LoRA parameters, and (2) a separate GRPO stage for training the budget predictor. Across different backbone models, the main fine-tuning stage requires approximately $8$--$25$ hours, while the GRPO stage is substantially lighter and consistently finishes within $1.5$ hours. This result indicates that the additional cost introduced by budget prediction is modest compared with the primary fine-tuning stage, making the overall training pipeline practical and reproducible.

\subsection{Training Budget Predictor with GRPO}\label{app:grpo_training}

\paragraph{Problem Setup.}
Given an input $x$, we extract the KV-aware context $G=\phi(x)$ using the same feature extraction as described in \Cref{sec:kv_aware}. The Budget Predictor is a categorical policy $\pi_\theta(k\mid G)$ over the action space $\mathcal{A}=\{1,2,\dots,L_{\text{set}}\}$, where action $k$ corresponds to executing $k$ middle layers. All other model parameters remain frozen during training.

\paragraph{Reward Function.}
For a target sequence $y$, we run the frozen model under budget $b(k)=\frac{k+2}{L}$ in teacher-forcing mode to obtain token probabilities $\hat{p}_{b(k)}$. The per-sample cross-entropy loss is:
\begin{equation}
\mathrm{CE}\big(\hat{p}_{b(k)},y\big) = -\frac{1}{T-1}\sum_{t=2}^{T}\log \hat{p}_{b(k)}\!\left(y_t \mid x, y_{<t}\right).
\end{equation}

We balance predictive performance and computational cost via:
\begin{equation}
r(k) = -\lambda_{\text{perf}} \cdot \mathrm{CE}\big(\hat{p}_{b(k)},y\big) - \lambda_{\text{cost}} \cdot b(k),
\end{equation}
where $\lambda_{\text{perf}}, \lambda_{\text{cost}} > 0$ are hyperparameters that control the performance-efficiency trade-off.

\paragraph{Group-Relative Advantage.}
For each training example $(x,y)$, we sample a group of $M$ actions $k^{(j)} \sim \pi_\theta(\cdot\mid G)$, $j=1,\dots,M$, and compute their rewards $r^{(j)}=r(k^{(j)})$. To reduce variance, we normalize rewards within each group:
\begin{equation}
\mu = \frac{1}{M}\sum_{j=1}^{M} r^{(j)}, \quad 
\sigma = \sqrt{\frac{1}{M}\sum_{j=1}^{M}(r^{(j)}-\mu)^2}, \quad
A^{(j)} = \frac{r^{(j)}-\mu}{\sigma+\varepsilon},
\end{equation}
where $\varepsilon > 0$ ensures numerical stability.

\paragraph{Optimization Objective.}
GRPO maximizes the advantage-weighted log-likelihood with entropy regularization:
\begin{equation}
\mathcal{L}_{\text{GRPO}}(\theta) = -\frac{1}{M}\sum_{j=1}^{M} A^{(j)} \log \pi_\theta(k^{(j)}\mid G) - \beta \mathcal{H}(\pi_\theta(\cdot\mid G)),
\end{equation}
where $\beta \geq 0$ encourages exploration and $\mathcal{H}(\cdot)$ is the categorical entropy. Only the Budget Predictor parameters $\theta$ receive gradients; the language model and Decision Module remain frozen. This directly learns a discrete layer-allocation policy that maximizes task quality per unit of compute.

\section{Experiment Results}\label{app:more_results}

\subsection{Reproduce of Baselines}\label{app:baselines}
We compare against recent depth-pruning approaches. \emph{Static} methods:
\textbf{(1) Shortened LLaMA} \citep{kim2024shortened} \footnote{\url{https://github.com/Nota-NetsPresso/shortened-llm}} determines block importance by measuring the increase in perplexity ($\Delta\text{PPL}$) when each block is removed, then prunes the least important ones;
\textbf{(2) ShortGPT} \citep{men2024shortgpt} \footnote{\url{https://github.com/sramshetty/ShortGPT}} ranks layers based on the cosine similarity between their input and output hidden states, treating layers with high similarity as redundant;
\textbf{(3) SLEB} \citep{song2024sleb} \footnote{\url{https://github.com/jiwonsong-dev/SLEB}} iteratively removes the layer that causes the smallest $\Delta\text{PPL}$, and verifies that the omission is safe by checking next-token prediction consistency on a calibration set.
\emph{Dynamic} methods:
\textbf{(4) PuDDing} \citep{wee2025pudding} \footnote{\url{https://github.com/tada0347/PuDDing}} constructs token-level omission sets by evaluating commonsense reasoning benchmarks, enabling dynamic layer skipping depending on the input difficulty. We use the official omission-set construction and extend it with LoRA under our unified training setup. ;
\textbf{(5) FiRST} \citep{jain2025first} equips each transformer layer with a lightweight linear router that makes execute/skip decisions based on the layer input. We reproduced FiRST from the original paper and verified its performance in our setting.

\begin{table*}[t!]
\centering
\caption{
Performance comparison on Llama2-7B across eight Commonsense reasoning benchmarks for different pruning methods at various sparsity levels (12.5\%-50\%). \textbf{Boldface} indicates the best performance, and the \underline{Underline} means the second-order performance. \textsc{Buddy} achieves the best average performance at moderate-to-high sparsity levels and remains competitive at light sparsity.
}
\label{tab:result_llama2}
\resizebox{\textwidth}{!}{
\begin{tabular}{l|c|c|ccccccccc}
\toprule
Method & Pruning & RM Blocks & OBQA & PIQA & BoolQ & SIQA & Hellaswag & ARC-E & ARC-C & Winogrande & Avg. \\
\midrule
Dense w/o	& -	& 0(0.0\%)	& 44.20 & 79.11 &	77.71 &	46.06 &	76.02 &	76.30 &	46.33 &	69.14 &	64.36 \\
\midrule
Shortened Llama & static & 4(12.5\%) & 40.60 & \textbf{78.02} & 71.25 & 45.29 & 72.53 & \underline{73.27} & 42.83 & 62.90 & 60.84 \\
ShortGPT & static & 4(12.5\%) & \underline{42.00} & 77.15 & \textbf{78.69} & \textbf{47.59} & \textbf{73.69} & 72.56 & \underline{44.45} & \textbf{69.14} & \textbf{63.16} \\
SLEB & static & 4(12.5\%) & 41.40 & \underline{77.48} & 71.96 & 45.14 & 72.60 & \textbf{73.40} & 41.04 & 64.01 & 60.88 \\
PuDDing & dynamic & 4(12.5\%) & 40.00 & 75.95 & \underline{73.55} & 43.30 & 69.73 & 69.32 & 39.51 & 64.48 & 59.48 \\
FiRST & dynamic & 4(12.5\%) & 36.00 & 56.20 & 57.74 & 40.48 & 47.66 & 48.27 & 34.73 & 54.22 & 46.91 
 \\
\rowcolor{gray!12} \textsc{Buddy} & dynamic & 4(12.5\%) & \textbf{43.00} & 77.09 & 73.30 & \underline{47.34} & \underline{73.64} & 73.02 & \textbf{44.80} & \underline{68.82} & \underline{62.63} \\
\midrule
Shortened Llama & static & 8(25\%) & 37.00 & \textbf{74.27} & 61.87 & 41.10 & 63.15 & 63.26 & 34.90 & 54.62 & 53.77 \\
ShortGPT & static & 8(25\%) & \underline{39.60} & 72.42 & 62.94 & \underline{43.71} & \textbf{67.66} & \underline{65.28} & \textbf{38.91} & \textbf{67.09} & \underline{57.20} \\
SLEB & static & 8(25\%) & \textbf{39.80} & \underline{73.78} & \underline{69.24} & 43.19 & 65.72 & \textbf{66.58} & 35.49 & 60.46 & 56.78 \\
PuDDing & dynamic & 8(25\%) & 36.60 & 71.82 & 62.87 & 39.30 & 60.30 & 61.41 & 35.58 & 56.67 & 53.07 \\
FiRST & dynamic & 8(25\%) & 35.40 & 55.22 & 57.74 & 40.38 & 44.52 & 45.20 & 32.94 & 40.38 & 43.97 \\
\rowcolor{gray!12} \textsc{Buddy} & dynamic & 8(25\%) & 39.20 & 73.18 & \textbf{72.63} & \textbf{45.65} & \underline{66.64} & \underline{65.28} & \underline{38.05} & \underline{66.85} & \textbf{58.44} \\
\midrule
Shortened Llama & static & 12(37.5\%) & 34.20 & \underline{70.89} & 62.14 & 39.87 & 52.81 & 57.28 & 29.86 & 52.33 & 49.92 \\
ShortGPT & static & 12(37.5\%) & 33.20 & 66.92 & \underline{71.22} & \textbf{43.65} & \textbf{58.76} & 54.46 & \textbf{34.13} & \textbf{64.09} & \underline{53.31} \\
SLEB & static & 12(37.5\%) & \textbf{36.00} & \textbf{71.38} & 60.28 & 41.15 & 54.83 & \textbf{59.76} & 31.06 & 52.96 & 50.93 \\
PuDDing & dynamic & 12(37.5\%) & 31.60 & 64.69 & 46.42 & 37.36 & 47.36 & 49.16 & 29.35 & 53.59 & 44.94 \\
FiRST & dynamic & 12(37.5\%) & \underline{36.20} & 55.17 & 55.87 & 39.82 & 43.99 & 45.33 & 30.89 & 53.67 & 45.12 \\
\rowcolor{gray!12} \textsc{Buddy} & dynamic & 12(37.5\%) & 35.60 & 70.78 & \textbf{72.81} & \underline{41.81} & \underline{57.87} & \underline{58.08} & \underline{31.14} & \underline{60.22} & \textbf{53.54} \\
\midrule
Shortened Llama & static & 16(50\%) & 30.80 & \textbf{65.51} & \underline{62.17} & \textbf{39.76} & 35.46 & \textbf{49.71} & 26.62 & 51.30 & 45.17 \\
ShortGPT & static & 16(50\%) & 29.80 & 61.81 & \textbf{62.20} & \underline{39.00} & 47.13 & 44.91 & 29.18 & \textbf{57.22} & \underline{46.41} \\
SLEB & static & 16(50\%) & 31.00 & \underline{65.07} & 61.71 & 38.84 & 42.93 & \underline{49.49} & 26.37 & 52.57 & 46.00 \\
PuDDing & dynamic & 16(50\%) & 31.60 & 64.47 & 46.42 & 37.36 & \underline{47.36} & 49.16 & \textbf{29.35} & \underline{53.59} & 44.91 \\
FiRST & dynamic & 16(50\%) & \textbf{33.20} & 53.21 & 54.53 & 32.91 & 39.73 & 41.12 & \underline{29.27} & 53.28 & 42.15 \\
\rowcolor{gray!12} \textsc{Buddy} & dynamic & 16(50\%) & \underline{32.80} & 64.74 & 60.21 & 37.92 & \textbf{48.66} & 48.91 & 28.41 & 53.51 & \textbf{46.90} \\
\bottomrule
\end{tabular}}
\end{table*}

\subsection{Results on Llama2-7B}
We also conduct experiments on Llama2-7B. Across eight commonsense reasoning benchmarks and four sparsity settings, \textbf{\textsc{Buddy}} exhibits strong average performance and robust degradation as pruning deepens from \Cref{tab:result_llama2}. At $\,12.5\%$ sparsity (removing 4 blocks), \textsc{Buddy} attains an average of $\mathbf{62.63}$, ranking \underline{second} and trailing the best baseline (ShortGPT, $63.16$) by only $0.53$ points. For higher sparsities, \textsc{Buddy} delivers the best average in every case: $\mathbf{58.44}$ at $25\%$ ($+1.24$ over the best baseline), $\mathbf{53.54}$ at $37.5\%$ ($+0.23$), and $\mathbf{46.90}$ at $50\%$ ($+0.49$). Relative to the dense (unpruned) model average of $64.36$, \textsc{Buddy} preserves $\,97.3\%$, $90.8\%$, $83.2\%$, and $72.9\%$ of accuracy at $12.5\%$, $25\%$, $37.5\%$, and $50\%$ sparsity, respectively. \textsc{Buddy} delivers the best \emph{average} accuracy at moderate-to-high sparsity ($\geq 25\%$) while remaining highly competitive at light sparsity. Moreover, its budget flexibility makes it adapt to different sparsity ratios using only one model.

\begin{table*}[t!]
\centering
\caption{
Performance comparison on Llama1-13B across eight Commonsense Reasoning benchmarks for different pruning methods at various sparsity levels (12.5\%-50\%). \textbf{Boldface} indicates the best performance, and the \underline{Underline} means the second-order performance. \textsc{Buddy} achieves the best average performance at moderate-to-high sparsity levels and remains competitive at light sparsity.
}
\label{tab:result_llama13b}
\resizebox{\textwidth}{!}{
\begin{tabular}{l|c|c|ccccccccc}
\toprule
Method & Pruning & RM Blocks & OBQA & PIQA & BoolQ & SIQA & Hellaswag & ARC-E & ARC-C & Winogrande & Avg. \\
\midrule
Dense w/o & - & 0(0.0\%) & 44.80 & 80.14 & 77.92 & 46.72 & 79.06 & 77.36 & 47.61 & 72.69 & 65.79 \\

\midrule
Shortened Llama & static & 5(12.5\%) & 43.00 & \underline{79.60} & 67.43 & 47.75 & 77.38 & \textbf{76.85} & \underline{48.46} & 71.51 & 64.00 \\
ShortGPT & static & 5(12.5\%) & \textbf{44.80} & 79.54 & 73.82 & \underline{48.67} & \textbf{78.04} & 76.47 & \underline{48.46} & \underline{72.85} & \underline{65.33} \\
SLEB & static & 5(12.5\%) & 43.80 & 78.56 & \underline{74.31} & 46.47 & 76.73 & 74.92 & 43.94 & 68.90 & 63.45 \\
PuDDing	& dynamic & 5(12.5\%) & 42.60 &	78.45 &	67.77 &	45.45 & 76.11 & 73.36 &	43.94 &	70.24 &	62.24 \\
\rowcolor{gray!12} \textsc{Buddy} & dynamic & 5(12.5\%) & \underline{44.00} & \textbf{80.03} & \textbf{76.33} & \textbf{48.82} & 77.54 & \underline{76.64} & \textbf{49.06} & \textbf{72.93} & \textbf{65.67} \\

\midrule
Shortened Llama & static & 10(25\%) & 40.20 & \textbf{77.75} & 58.01 & 47.65 & 73.57 & \textbf{72.98} & 44.37 & 69.06 & 60.45 \\
ShortGPT & static & 10(25\%) & \underline{42.00} & \underline{77.58} & 54.62 & \textbf{48.31} & \textbf{74.30} & 72.22 & \underline{45.31} & \underline{71.27} & 60.70 \\
SLEB & static & 10(25\%) & \underline{42.00} & 77.31 & \underline{73.06} & 46.52 & 72.97 & 72.10 & 42.83 & 64.09 & \underline{61.36} \\
PuDDing	& dynamic & 10(25\%) & 38.20 &	74.43 &	69.14 &	42.22 &	69.30 &	69.95 &	38.74 &	63.06 &	58.13 \\
\rowcolor{gray!12} \textsc{Buddy} & dynamic & 10(25\%) & \textbf{43.40} & 77.37 & \textbf{77.68} & 46.78 & \underline{74.27} & 72.26 & \textbf{46.16} & \textbf{71.51} & \textbf{63.68} \\

\midrule
Shortened Llama & static & 15(37.5\%) & \underline{38.20} & 72.80 & 49.88 & \textbf{46.11} & 68.82 & \underline{66.37} & \underline{40.70} & 68.03 & 56.36 \\
ShortGPT & static & 15(37.5\%) & 37.40 & 72.03 & 66.97 & \underline{45.91} & \textbf{70.12} & 65.99 & \textbf{42.06} & \underline{68.82} & \underline{58.66} \\
SLEB & static & 15(37.5\%) & 36.40 & \textbf{73.12} & \underline{69.88} & 43.96 & 65.51 & \textbf{66.58} & 34.64 & 61.80 & 56.49 \\
PuDDing	& dynamic & 15(37.5\%) & 34.00 & 69.64 &65.84 &	38.38 &	60.13 &	60.73 &	31.66 &	58.80 &	52.40 \\ 
\rowcolor{gray!12} \textsc{Buddy} & dynamic & 15(37.5\%) & \textbf{39.00} & 72.69 & \textbf{79.39} & 45.60 & 68.55 & 63.26 & 40.36 & \textbf{69.61} & \textbf{59.81} \\

\midrule
Shortened Llama & static & 20(50\%) & 32.40 & \textbf{69.37} & 44.13 & 42.12 & 56.01 & \underline{57.79} & 31.83 & 59.83 & 49.18 \\
ShortGPT & static & 20(50\%) & 33.80 & 66.16 & \underline{62.32} & \textbf{42.63} & \textbf{60.04} & 54.17 & \textbf{35.75} & \textbf{64.64} & \textbf{52.44} \\
SLEB & static & 20(50\%) & \underline{34.00} & \underline{68.99} & 61.31 & \underline{41.30} & 54.73 & \textbf{58.00} & 30.89 & 56.43 & 50.71 \\
PuDDing	& dynamic & 20(50\%) &30.60 & 64.91 & 61.99 &	37.72 &	50.38 &	48.48 &	25.68 &	55.72 &	46.94 \\ 
\rowcolor{gray!12} \textsc{Buddy} & dynamic & 20(50\%) & \textbf{34.40} & 66.43 & \textbf{63.21} & \underline{41.30} & \underline{59.67} & 52.02 & \underline{33.79} & \underline{64.01} & \underline{51.85} \\
\bottomrule
\end{tabular}}
\end{table*}

\subsection{Results on Llama1-13B}
We also conduct experiments on Llama1-13B with static baselines. Across eight commonsense reasoning benchmarks and four sparsity settings, \textbf{\textsc{Buddy}} exhibits strong average performance and robust degradation as pruning deepens from \Cref{tab:result_llama13b}. At $\,50\%$ sparsity (removing 16 blocks), \textsc{Buddy} attains an average of $\mathbf{51.85}$, ranking \underline{second} and trailing the best baseline (ShortGPT, $52.44$) by only $0.39$ points. For other sparsities, \textsc{Buddy} delivers the best average in every case: $\mathbf{65.67}$ at $12.5\%$ ($+1.67$ over the best baseline), $\mathbf{63.68}$ at $25\%$ ($+2.32$), and $\mathbf{59.81}$ at $37.5\%$ ($+1.15$). Relative to the dense (unpruned) model average of $65.79$, \textsc{Buddy} preserves $\,99.8\%$, $96.8\%$, $90.9\%$, and $78.8\%$ of accuracy at $12.5\%$, $25\%$, $37.5\%$, and $50\%$ sparsity, respectively.

\subsection{Results on Other Benchmarks}
We further evaluate the fine-tuned Llama2-7B and Qwen2.5-7B-Instruct models on a more challenging reasoning benchmark, GSM8K \citep{cobbe2021gsm8k}. Due to the more aggressive pruning ratios, all methods suffer substantial performance degradation on this dataset. Therefore, we focus on sparsity levels of 12.5\% and 25\%, and report the results in \Cref{tab:gsm8k_benchmark}. As shown in the \Cref{tab:gsm8k_benchmark}, the performance of all pruned models is noticeably lower than their dense counterparts at both sparsity levels. Nonetheless, \textsc{Buddy} achieves better overall performance than the baselines. The improvement is particularly significant on Qwen2.5-7B-Instruct: at 12.5\% sparsity, \textsc{Buddy} attains 51.10 accuracy, considerably outperforming the second-best SLEB (37.15); at 25\% sparsity, \textsc{Buddy} still leads by a large margin (14.63 vs. 6.60 for ShortGPT). These results demonstrate that our \textsc{Buddy} framework remains effective even on more difficult benchmarks and under aggressive pruning regimes.

\begin{table*}[h]
\centering
\caption{
Performance comparison on Llama 2-7B and Qwen 2.5-7B-Instruct on GSM8K benchmarks for different pruning methods at various sparsity levels (12.5\%-25\%). \textbf{Boldface} indicates the best performance, and the \underline{Underline} means the second-order performance. \textsc{Buddy} achieves the best average performance at moderate-to-high sparsity levels and remains competitive at light sparsity.
}
\label{tab:gsm8k_benchmark}
\resizebox{0.75\textwidth}{!}{
\begin{tabular}{l|c|c|cc}
\toprule
Method & Pruning method & RM Blocks & Llama2-7B & Qwen2.5-7B-Instruct \\
\midrule
Dense w/o        & -       & 0(0.0\%)   & 14.03 & 82.87 \\
\midrule
Shortened Llama  & static  & 4(12.5\%)  & 5.16  & 27.22 \\
ShortGPT         & static  & 4(12.5\%)  & \textbf{8.11} & 35.94 \\
SLEB             & static  & 4(12.5\%)  & 3.64  & \underline{37.15} \\
PuDDing          & dynamic & 4(12.5\%)  & 1.74  & 0.99 \\
FiRST            & dynamic & 4(12.5\%)  & 0.91  & 27.98 \\
\rowcolor{gray!12}\textsc{Buddy}            & dynamic & 4(12.5\%)  & \underline{7.81} & \textbf{51.10} \\
\midrule
Shortened Llama  & static  & 8(25\%)    & 2.05  & 5.84 \\
ShortGPT         & static  & 8(25\%)    & \underline{2.50} & \underline{6.60} \\
SLEB             & static  & 8(25\%)    & 1.97  & 6.37 \\
PuDDing          & dynamic & 8(25\%)    & 0.99  & 0.61 \\
FiRST            & dynamic & 8(25\%)    & 1.21  & 2.58 \\
\rowcolor{gray!12}\textsc{Buddy}            & dynamic & 8(25\%)    & \textbf{2.65} & \textbf{14.63} \\
\bottomrule
\end{tabular}}
\end{table*}

\begin{table}[h]
\centering
\caption{Ablation on the Start Layer that applied Decision Module. Average accuracy (\%) is reported.}
\label{tab:ablation_start_layer}
\resizebox{0.5\textwidth}{!}{
\begin{tabular}{ccccc}
\toprule
Start layer & \multicolumn{4}{c}{RM Blocks} \\
\cmidrule(lr){2-5}
& 4 (12.5\%) & 8 (25\%) & 12 (37.5\%) & 16 (50\%) \\
\midrule
1 & \textbf{62.71} & \textbf{57.65} & 51.54 & 43.81 \\
2 & \underline{61.76} & \underline{57.07} & \underline{51.91} & 44.16 \\
3 & 60.61 & 56.37 & \textbf{52.36} & \textbf{47.80} \\
4 & 60.92 & 56.89 & 50.91 & \underline{46.95} \\
\bottomrule
\end{tabular}}
\end{table}

\subsection{Speed Analysis}\label{app:speed_analysis}

We measure end-to-end throughput (tokens/s) on Alpaca and SAMSum for both \emph{prefill} and \emph{decode} phases. During decoding, the model generates 128 tokens. We compare the throughput with baselines that are dynamic inference. Results show in \Cref{tab:result_speed_full}, indicating that in the prefill stage, \textsc{Buddy} outperforms existing dynamic inference methods in throughput at all sparsity rates. However, existing methods, due to excessive per-layer routing or switching between different LoRAs, have limited acceleration and can even be slower than Dense methods at low sparsity rates. While in the decode stage, \textsc{Buddy} left behind the PuDDing method due to its fixed inference path. In summary, the \textsc{Buddy} method not only offers greater versatility but also delivers greater throughput gains.

\begin{table}[ht]
\centering
\caption{Speed analysis on the Alpaca and SAMSum datasets. The speed is measured as tokens/s.}
\label{tab:result_speed_full}
\resizebox{0.9\linewidth}{!}{
\begin{tabular}{l|c|cccc|cccc}
\toprule
 \multirow{2}{*}{Method} & \multirow{2}{*}{RM Blocks} & \multicolumn{4}{c|}{Alpaca} & \multicolumn{4}{c}{SAMSum} \\
\cmidrule(r){3-10}
 & & Prefill & Speed up & Decode & Speed Up & Prefill & Speed Up & Decode & Speed Up \\
\midrule
Dense & 0(0.0\%) & 1753.16 & $\times$1.00 & 39.48 & $\times$1.00 & 4033.83 & $\times$1.00 & 60.70 & $\times$1.00 \\
\midrule
PuDDing & 4(12.5\%) & 1491.73 & $\times$0.85 & 47.70 & $\times$1.21 & 3255.28 & $\times$0.81 & 72.84 & $\times$1.20 \\
FiRST & 4(12.5\%) & 1794.51 & $\times$1.02 & 33.46 & $\times$0.85 & 4075.96 & $\times$1.01 & 51.58 & $\times$0.85 \\
\rowcolor{gray!12}\textsc{Buddy} & 4(12.5\%) & 1991.54 & $\times$1.14 & 40.03 & $\times$1.01 & 4035.68 & $\times$1.00 & 61.28 & $\times$1.01 \\
\midrule
PuDDing & 8(25\%) & 1691.74 & $\times$0.96 & 55.96 & $\times$1.42 & 3662.78 & $\times$0.91 & 84.29 & $\times$1.39 \\
FiRST & 8(25\%) & 1795.22 & $\times$1.02 & 33.42 & $\times$0.85 & 3972.77 & $\times$0.98 & 51.05 & $\times$0.84 \\
\rowcolor{gray!12}\textsc{Buddy} & 8(25\%) & 2440.14 & $\times$1.39 & 46.80 & $\times$1.19 & 4777.07 & $\times$1.18 & 71.91 & $\times$1.18 \\
\midrule
PuDDing & 12(37.5\%) & 1837.07 & $\times$1.05 & 62.82 & $\times$1.59 & 3951.91 & $\times$0.98 & 95.53 & $\times$1.57 \\
FiRST & 12(37.5\%) & 1780.78 & $\times$1.02 & 33.30 & $\times$0.84 & 4082.99 & $\times$1.01 & 51.63 & $\times$0.85 \\
\rowcolor{gray!12}\textsc{Buddy} & 12(37.5\%) & 2548.39 & $\times$1.45 & 51.49 & $\times$1.30 & 5476.04 & $\times$1.36 & 79.11 & $\times$1.30 \\
\midrule
PuDDing & 16(50\%) & 2196.42 & $\times$1.25 & 81.63 & $\times$2.07 & 4836.26 & $\times$1.20 & 124.85 & $\times$2.06 \\
FiRST & 16(50\%) & 1792.25 & $\times$1.02 & 33.49 & $\times$0.85 & 4041.14 & $\times$1.00 & 51.20 & $\times$0.84 \\
\rowcolor{gray!12}\textsc{Buddy} & 16(50\%) & 3348.18 & $\times$1.91 & 64.84 & $\times$1.64 & 6581.55 & $\times$1.63 & 99.12 & $\times$1.63 \\
\bottomrule
\end{tabular}}
\end{table}

\subsection{Layer Selection Analysis}\label{app:speed_analysis}
\begin{figure}
\centering
\includegraphics[width=\linewidth]{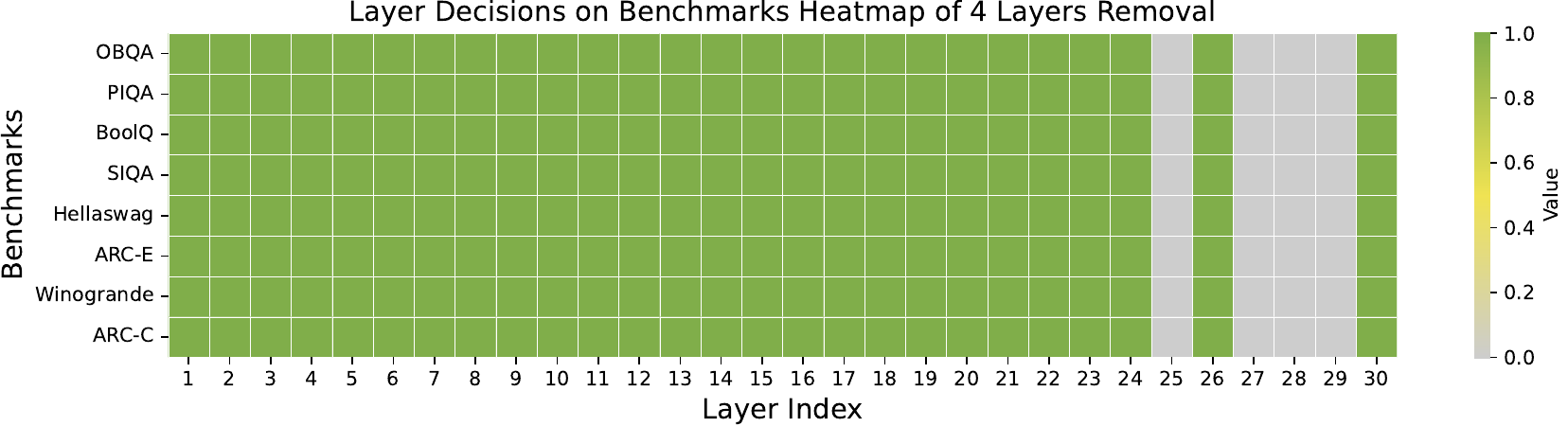}
\caption{
A visual illustration of the \textsc{Buddy}'s average decision of each transformer block on 12.5\% sparsity (4 layers removed). The color \textcolor{darkgreen}{green} indicates that the blocks are likely important to be maintained, the color \textcolor{darkgrey}{grey} indicates that the blocks are likely to be pruned, and the color \textcolor{darkyellow}{yellow} means the importance of layers varies across tasks and samples. 
}
\label{fig:heat_map_rm4}
\end{figure}

\begin{figure}
\centering
\includegraphics[width=\linewidth]{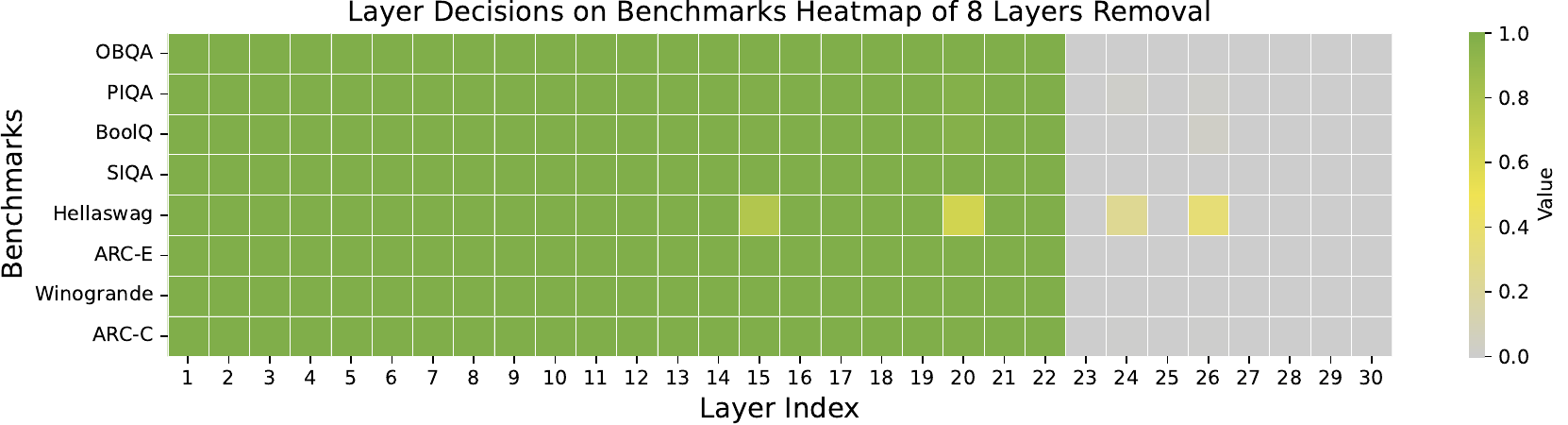}
\caption{
A visual illustration of the \textsc{Buddy}'s average decision of each transformer block on 25.0\% sparsity (8 layers removed). The color \textcolor{darkgreen}{green} indicates that the blocks are likely important to be maintained, the color \textcolor{darkgrey}{grey} indicates that the blocks are likely to be pruned, and the color \textcolor{darkyellow}{yellow} means the importance of layers varies across tasks and samples. 
}
\label{fig:heat_map_rm8}
\end{figure}

\begin{figure}
\centering
\includegraphics[width=\linewidth]{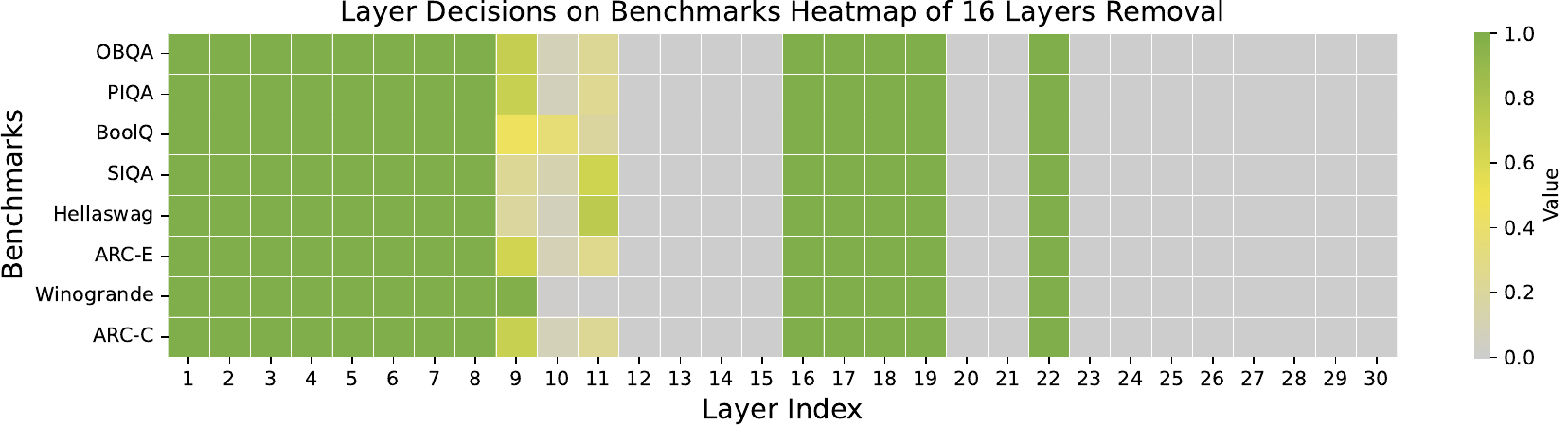}
\caption{
A visual illustration of the \textsc{Buddy}'s average decision of each transformer block on 50.0\% sparsity (16 layers removed). The color \textcolor{darkgreen}{green} indicates that the blocks are likely important to be maintained, the color \textcolor{darkgrey}{grey} indicates that the blocks are likely to be pruned, and the color \textcolor{darkyellow}{yellow} means the importance of layers varies across tasks and samples. 
}
\label{fig:heat_map_rm16}
\end{figure}

We analyze routing decisions across tasks and inputs under different sparsity levels; results appear in \Cref{fig:heat_map_rm4} (12.5\%), \Cref{fig:heat_map_rm8} (25\%), and \Cref{fig:heat_map_rm16} (50\%). At \textbf{low sparsity} (12.5\%), layer-selection decisions are relatively stable across inputs, indicating a near-fixed path. As sparsity increases, the \textbf{importance distribution becomes less uniform}. At \textbf{25\% sparsity}, layers 15, 20, 24, and 26 undergo notable shifts in importance; by \textbf{50\% sparsity}, layers 9--11 exhibit the largest changes and their importance varies substantially across tasks. Moreover, at 25\% sparsity the \textsc{HellaSwag} task assigns different importance to layers 24 and 26, while at 50\% sparsity this variability shifts to layers 9--11. These observations underscore that layer importance is \emph{not invariant} to the sparsity regime; consequently, routing must adapt dynamically as the budget changes.

\subsection{Budget Predictor Training}\label{app:budget_predictor_training}
\begin{figure*}[htbp]
  \centering
  \begin{subfigure}[b]{0.49\linewidth}
    \includegraphics[width=\linewidth]{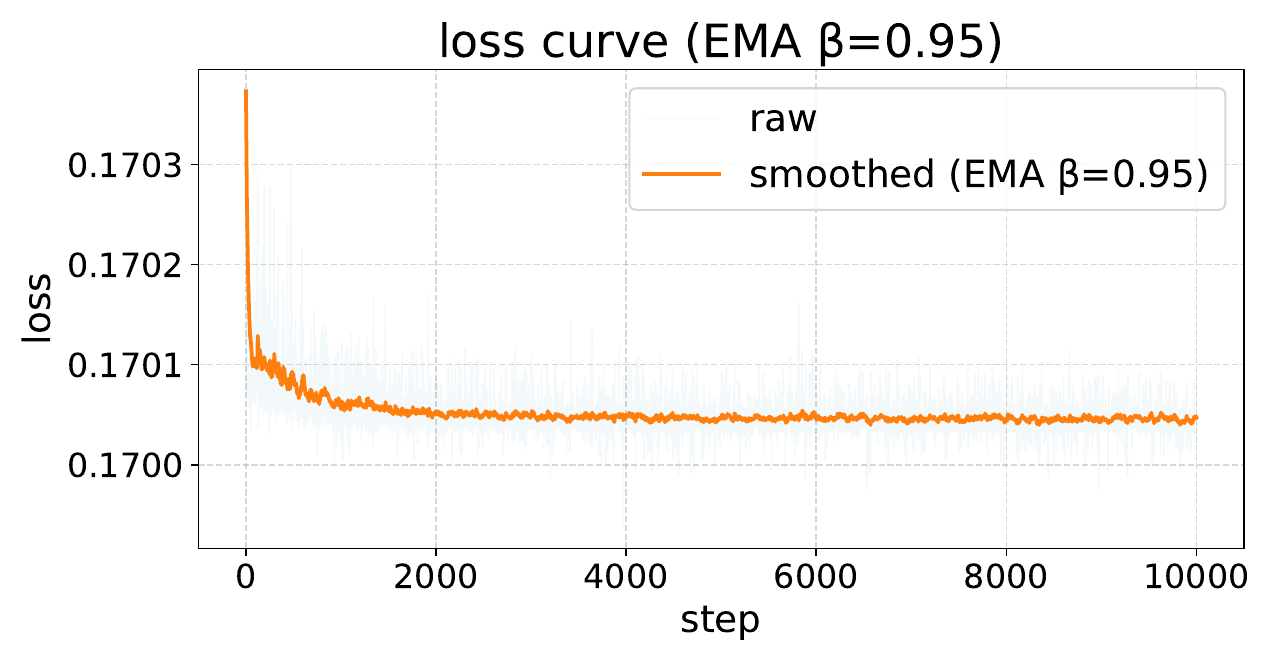}
    \caption{Training Loss of Budget Predictor via GRPO.}
    \label{fig:grpo_loss}
  \end{subfigure}\hfil
  \begin{subfigure}[b]{0.49\linewidth}
    \includegraphics[width=\linewidth]{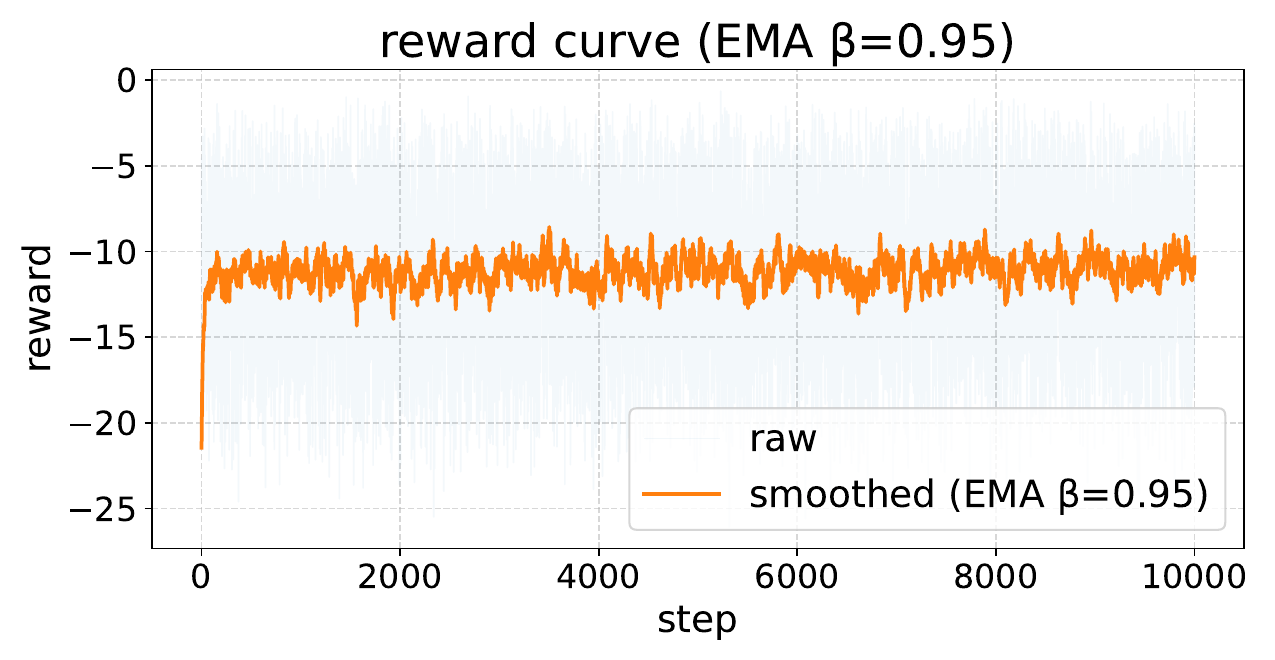}
    \caption{Training Reward of Budget Predictor via GRPO.}
    \label{fig:grpo_reward}
  \end{subfigure}
  \begin{subfigure}[b]{0.49\linewidth}
    \includegraphics[width=\linewidth]{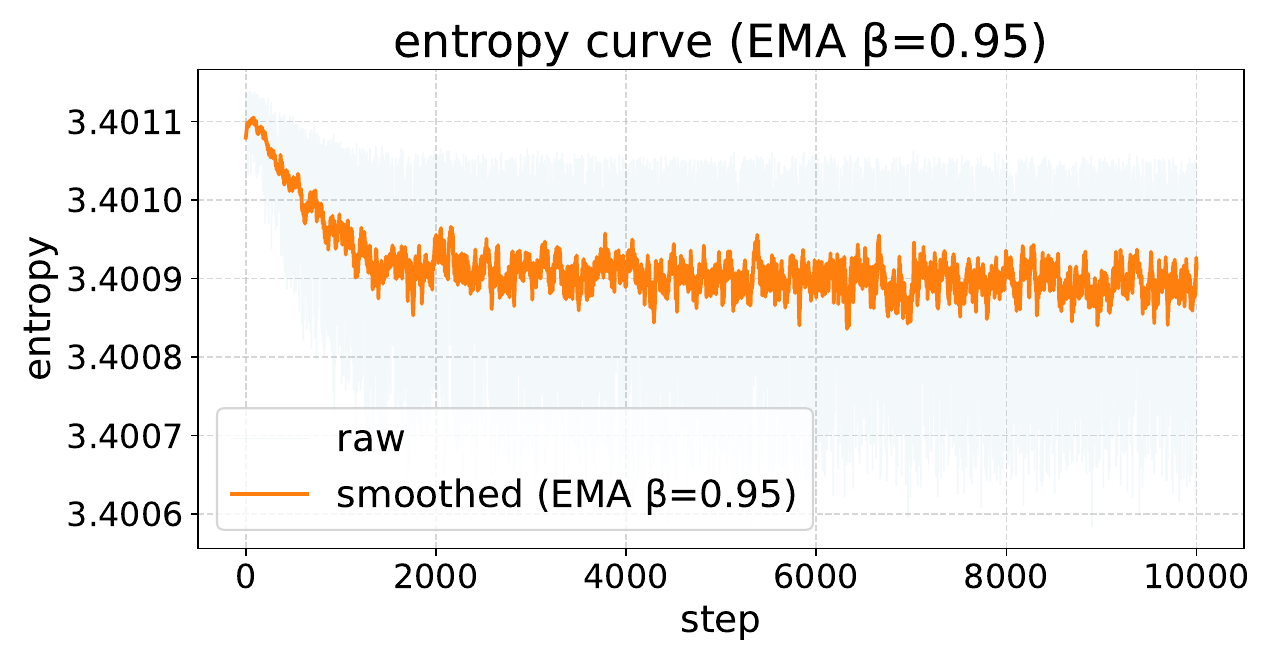}
    \caption{Training Entropy of Budget Predictor via GRPO.}
    \label{fig:grpo_entropy}
  \end{subfigure}\hfil
  \begin{subfigure}[b]{0.49\linewidth}
    \includegraphics[width=\linewidth]{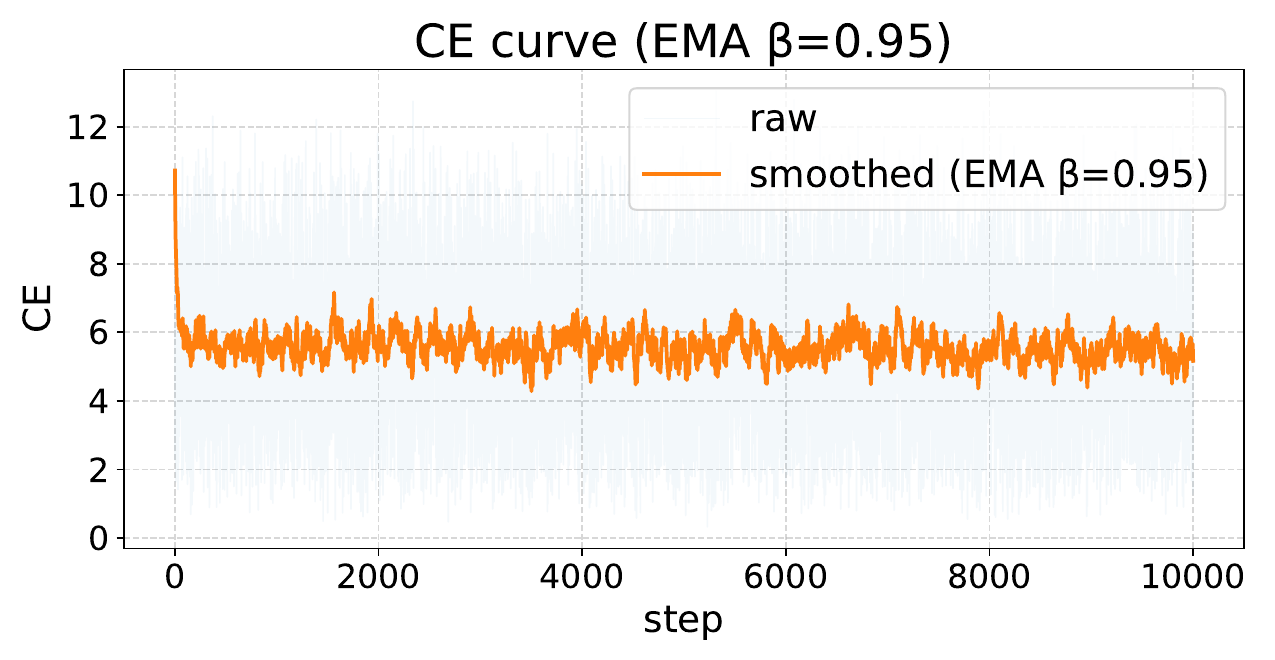}
    \caption{Training CE of Budget Predictor via GRPO.}
    \label{fig:grpo_ce}
  \end{subfigure}
  \caption{Training details of GRPO. Training loss, reward, CE, and entropy are reported.}
  \label{fig:grpo_result}
\end{figure*}

We train the \textbf{Budget Predictor} with GRPO while \emph{freezing} the LLM backbone and the Decision Module. Training uses the \textsc{Alpaca} dataset. We set the group size to $G=4$, learning rate to $1\times10^{-5}$, and optimize with Adam at batch size $2$ for $10{,}000$ steps. Unless otherwise noted, we adopt the reward from \Cref{app:grpo_training} with coefficients
$\lambda_{\text{perf}}=5.0$ and $\lambda_{\text{cost}}=0.05$. \Cref{fig:grpo_result} reports the training curves of training loss, training reward, training CE, and training entropy, showing a steady decrease in $\mathcal{L}_{\text{GRPO}}$ and a concurrent increase in reward, indicating that GRPO effectively learns a stable, discrete budget policy under our setup.

\subsection{Budget Predictor Evaluation and Analysis}\label{app:budget_predictor_evaluation}
We conduct a quantitative evaluation of the \textbf{Budget Predictor} on eight commonsense reasoning benchmarks together with two higher-variance tasks, \textsc{MMLU} and \textsc{GSM8K}. At inference time, we do \emph{not} provide a user-specified budget; instead, the predictor automatically selects an instance-wise budget $\hat b$ for each input.

\begin{figure*}
\centering
\includegraphics[width=\linewidth]{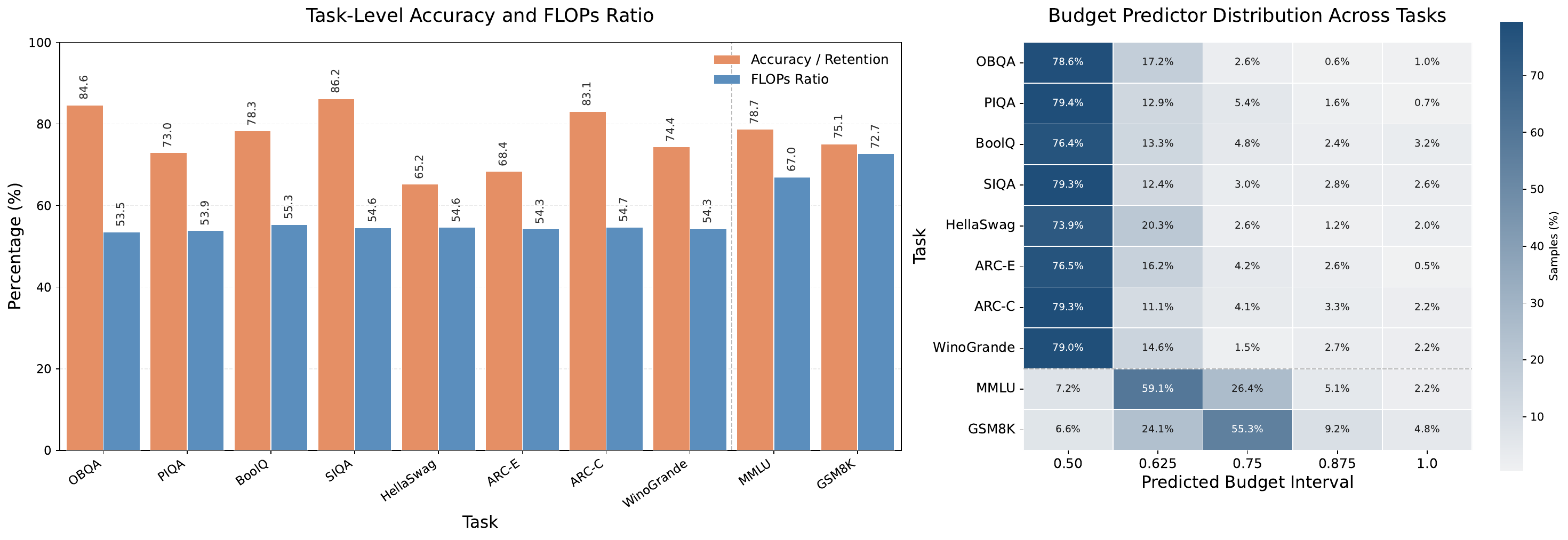}
\caption{Budget Predictor evaluation and analysis across tasks. Left: average performance retention and average execution ratio when the budget is selected automatically by the Budget Predictor. Right: instance-level budget distribution over five discrete budget intervals, showing that the predictor adapts its compute allocation across tasks rather than collapsing to a single budget.}
\label{fig:budget_predictor_acc}
\end{figure*}

\Cref{fig:budget_predictor_acc} summarizes two complementary views. The left panel reports the average execution ratio and the corresponding performance retention rate. On the commonsense benchmarks, the predictor allocates a relatively similar amount of compute on average (roughly $53\%\!\sim\!55\%$), while the retained performance still varies across tasks, indicating that the same average compute budget can lead to different task-level robustness. In contrast, the higher-variance tasks require noticeably larger average budgets, suggesting that the predictor tends to allocate more computation when the task distribution is more challenging.

More importantly, the right panel visualizes the full distribution of predicted budget intervals, which directly addresses the concern of mode collapse. The predictor does \emph{not} degenerate to a constant output. For the commonsense benchmarks, most samples concentrate in the lower budget intervals, especially around $0.50$ and $0.625$, with non-trivial mass still assigned to higher intervals. For \textsc{MMLU} and \textsc{GSM8K}, the distribution shifts further toward larger budgets: \textsc{MMLU} is mainly concentrated around the $0.625$ interval, while \textsc{GSM8K} places the largest mass on the $0.75$ interval. This pattern shows that the Budget Predictor is both task-sensitive and input-sensitive, rather than behaving as a fixed static strategy with a nearly constant execution rate.

\section{Background}
\paragraph{LLMs Architecture.} 
Given a tokenized sequence $\mathcal{X}=\{x_1,x_2,\dots,x_n\}$, an autoregressive LLM predicts the next token $x_{n+1}$ by applying an embedding layer, a stack of $L$ Transformer blocks, and a classification head:
\begin{equation}
h_n \;=\; \mathcal{F}_L \circ \mathcal{F}_{L-1} \circ \cdots \circ \mathcal{F}_1 \circ \mathcal{F}_{\text{embed}}(\mathcal{X}),\qquad
x_{n+1} \sim \mathrm{Softmax}\!\big(\mathcal{F}_{\text{head}}(h_n)\big).
\end{equation}
Here $\mathcal{F}_{\text{embed}}$ denotes token embeddings, $\{\mathcal{F}_\ell\}_{\ell=1}^L$ are Transformer blocks, and $\mathcal{F}_{\text{head}}$ maps the final hidden state to logits.

\paragraph{Prefill vs.\ Decode.}\label{sec:prefill_vs_decode}
Inference consists of a \emph{prefill} phase and an \emph{autoregressive decode} phase. In prefill, the model processes the entire prompt with teacher forcing. During decoding, the model generates one token at a time and appends it to the context. To avoid recomputing attention over the full history at every step, LLMs maintain a \emph{KV cache} of keys/values from prior positions. Let $t$ be the current decoding step and $(Q_t,K_t,V_t)$ the query/key/value of the new token. With cached states $K_{\text{cache}}^{(t-1)},V_{\text{cache}}^{(t-1)}$, attention uses and the cache is updated by:
\begin{equation}
\label{equotion:kv_cache}
\mathrm{Attn}\!\big(Q_t,\;[K_{\text{cache}}^{(t-1)};K_t],\;[V_{\text{cache}}^{(t-1)};V_t]\big), \qquad
K_{\text{cache}}^{(t)}=[K_{\text{cache}}^{(t-1)};K_t],\qquad
V_{\text{cache}}^{(t)}=[V_{\text{cache}}^{(t-1)};V_t].
\end{equation}
This mechanism substantially reduces decoding compute.

\section{Use Of LLMs}
In this study, LLMs are employed as automated proofreading agents to inspect, revise, and polish the manuscript. Specifically, they are tasked with detecting spelling and grammatical errors, assessing logical coherence and fluency of expression, and rewriting or condensing paragraphs where necessary.

\section{Software and Hardware Configuration}\label{appendix:software}
Our implementation utilizes the following configurations: \textit{PyTorch} version 2.1.2, \textit{Transformers} library version 4.41.0, \textit{PEFT (Parameter-Efficient Fine-Tuning)} library version 0.11.1, \textit{CUDA} version 12.4, \textit{GPU:} NVIDIA V100 GPU with 32GB of memory, NVIDIA A100 GPU with 80GB, \textit{Operating System:} Ubuntu. 

\section{LLM versions and Datasets}\label{appendix:LLM}
We provide the Hugging Face link of LLMs used in the experiment: 
Llama 2-7B: \url{https://huggingface.co/meta-llama/Llama-2-7b}; 
Llama 1-13B: \url{https://huggingface.co/yahma/llama-13b-hf}; 
Llama 3-8B: \url{https://huggingface.co/meta-llama/Meta-Llama-3.1-8B}; 
Qwen2.5-7B: \url{https://huggingface.co/Qwen/Qwen2.5-7B-Instruct};
Alpaca: {\url{https://huggingface.co/datasets/yahma/alpaca-cleaned}}.